\documentclass{article}

 \usepackage[final]{neurips_2022}

\usepackage{amsmath,amsfonts,bm}

\def\eqref#1{equation~\ref{#1}}
\def\floor#1{\lfloor #1 \rfloor}
\def\1{\bm{1}}

\DeclareMathAlphabet{\mathsfit}{\encodingdefault}{\sfdefault}{m}{sl}
\SetMathAlphabet{\mathsfit}{bold}{\encodingdefault}{\sfdefault}{bx}{n}

\usepackage{hyperref}
\usepackage{url}
\usepackage{graphics}
\usepackage{graphicx}
\usepackage{xcolor}

\usepackage{multirow}
\usepackage{subfigure}
\usepackage{xcolor}
\usepackage{caption,setspace}

\usepackage{wrapfig}
\newcommand{\cF}{{\mathcal F}}
\newcommand{\cG}{{\mathcal G}}

\newcommand{\cI}{{\mathcal I}}

\newcommand{\cO}{{\mathcal O}}

\newcommand{\vC}{{\bm{C}}}

\usepackage{pifont}% http://ctan.org/pkg/pifont

\newcommand{\highlight}[1]{\colorbox{blue!10}{#1}}
\definecolor{mygray}{gray}{0.4}
\usepackage[ruled,vlined]{algorithm2e}

\definecolor{commentcolor}{RGB}{110,154,155}   % define comment color
\newcommand{\PyComment}[1]{\ttfamily\textcolor{commentcolor}{\# #1}}  % add a "#" before the input text "#1"
\newcommand{\PyCode}[1]{\ttfamily\textcolor{black}{#1}} % \ttfamily is the code font
\newcommand{\g}[2]{#1\textsubscript{\textcolor{mygray}{$\pm$#2}}}

\title{Temporal Latent Bottleneck:  Synthesis of Fast and Slow Processing Mechanisms in Sequence Learning}

\author{
Aniket Didolkar \textsuperscript{1}, 
Kshitij Gupta \textsuperscript{1},
Anirudh Goyal \textsuperscript{1}, 
Nitesh B. Gundavarapu \textsuperscript{5}\\
\textbf{Alex Lamb \textsuperscript{2}}, 
\textbf{Nan Rosemary Ke \textsuperscript{3}}, 
\textbf{Yoshua Bengio \textsuperscript{1}\textsuperscript{,4}}}

\begin{document}

\let\footnote\relax\footnotetext{ \textsuperscript{1} Mila, University of Montreal, \textsuperscript{2} Microsoft Research, New York, NY, \textsuperscript{3} Google Deepmind,
\textsuperscript{4} CIFAR Fellow,
\textsuperscript{5} Google Research, Corresponding authors:  \texttt{adidolkar123@gmail.com} 
}
\maketitle

\begin{abstract}
Recurrent neural networks have a strong inductive bias towards learning temporally compressed representations, as the entire history of a sequence is represented by a single vector.  By contrast, Transformers have little inductive bias towards learning temporally compressed representations, as they allow for attention over all previously computed elements in a sequence.  Having a more compressed representation of a sequence may be beneficial for generalization, as a high-level representation may be more easily re-used and re-purposed and will contain fewer irrelevant details. At the same time, excessive compression of representations comes at the cost of expressiveness.  We propose a solution which divides computation into two streams.  A slow stream that is recurrent in nature aims to learn a specialized and compressed representation, by forcing chunks of $K$ time steps into a single representation which is divided into multiple vectors.  At the same time, a fast stream is parameterized as a Transformer to process chunks consisting of $K$ time-steps conditioned on the information in the slow-stream.  In the proposed approach we hope to gain the expressiveness of the Transformer, while encouraging better compression and structuring of representations in the slow stream. We show the benefits of the proposed method in terms of improved sample efficiency and generalization performance as compared to various competitive baselines for visual perception and sequential decision making tasks. 
\end{abstract}

\section{Introduction}
\vspace{-2mm}
%%% Compressed temporally information. 
%%% If there's no chunking structure, the model performs bad.
%%% focus on sequential structure, where it can be defined aproiri. 

%What do neural nets store in their representation of sequences?
%Cognitive science --> specialization of short term and long term memory.  
%More compressed --> better generalization, better scaling of retrieval.  More consistency within the sequence.  
%Less compressed --> better scaling (Transformers).  Faster processing, can immediately change on any given time step.  
%Proposal is a specialized method for combining a slow-long term memory stream with local attention.  

The interplay between fast and slow mechanisms for information processing and perception has been studied in both cognitive science and machine learning \cite{DBLP:conf/nips/BaHMLI16, Hinton87usingfast}.  In the brain, short-term and long-term memory have developed in a specialized way.  Short-term memory is allowed to change very quickly to react to immediate sensory inputs and perception.  It also tends towards high capacity storage of all pieces of information which may be relevant for future reasoning \cite{jonides2008mind, atkinson1971control, averbach1961short}. By contrast, long-term memory changes slowly \cite{kolodner1983maintaining, jeneson2012working}, is highly selective and involves repeated consolidation. It contains a set of memories that summarize the entire past, only storing details about observations which are most relevant \cite{goelet1986long, baddeley1984attention}.

%Research on scaling laws has shown that the expressiveness advantage of Transformers over recurrent neural networks grows with the length of the sequences being modeled \citep{devlin2018bert, Radford2018ImprovingLU, brown2020language}.

Deep Learning has seen a variety of architectures for processing sequential data \citep{hochrieter1997long, schuster1997bidirection, cho2014gru}. For example. recurrent neural networks compress information about a sequence into a single hidden state. Transformers get rid of the recurrent state by dynamically capturing information between positions using multi-head dot product attention \cite{vaswani2017attention}.  Transformers have become the dominant architecture across a wide range of domains including vision \citep{dosovitskiy2020vit}, natural language \citep{devlin2018bert, Radford2018ImprovingLU, brown2020language, zhang2022opt, chowdhery2022palm, rae2022scaling}, and reinforcement learning \citep{chen2021decision, janner2021reinforcement}. They have eclipsed recurrent neural networks \citep{hochrieter1997long,  schuster1997bidirection, cho2014gru} in almost all sequence processing domains due to their high representational capacity and scalability. Despite their wide applicability, it is well known that Transformers are very data hungry and work well mainly at scale. This can be attributed to their inductive bias towards modeling all possible pairwise interactions in the sequence which results in no consolidation of information. This lack of selectivity in the attention mechanism also leads to a high computational complexity which scales quadratically with input size. Additionally, modeling all possible pairwise interactions maybe extremely wasteful and may result in capturing unnecessary information not useful for the downstream task \citep{goyal2021coordination, jaegle2021perceiver}. The goal of this work is to design an architecture for autoregressive modeling that has an inductive bias towards learning temporally compressed representation that retains the benefits of Transformers  while preserving long-range interactions.

For learning temporally compressed representations, we start by dividing the computation of the Transformer into two streams of processing - a fast stream and a slow stream. Inspired by the idea of long-term and short-term memory, we want the fast stream to have a short-term memory with a high capacity that reacts quickly to sensory input. We refer to this fast stream as the perceptual module and implement it using a Transformer since they are known to have high representational capacity. On the other hand, we want the slow stream to have a long-term memory which updates at a slower rate and summarizes the most important information in the input sequence.  We refer to this slow stream as the \textbf{Temporal Latent Bottleneck}.

\begin{figure}[t]
  \vspace{-12mm}
  \begin{center}
    \includegraphics[width = 9cm, height = 4.5cm, keepaspectratio]{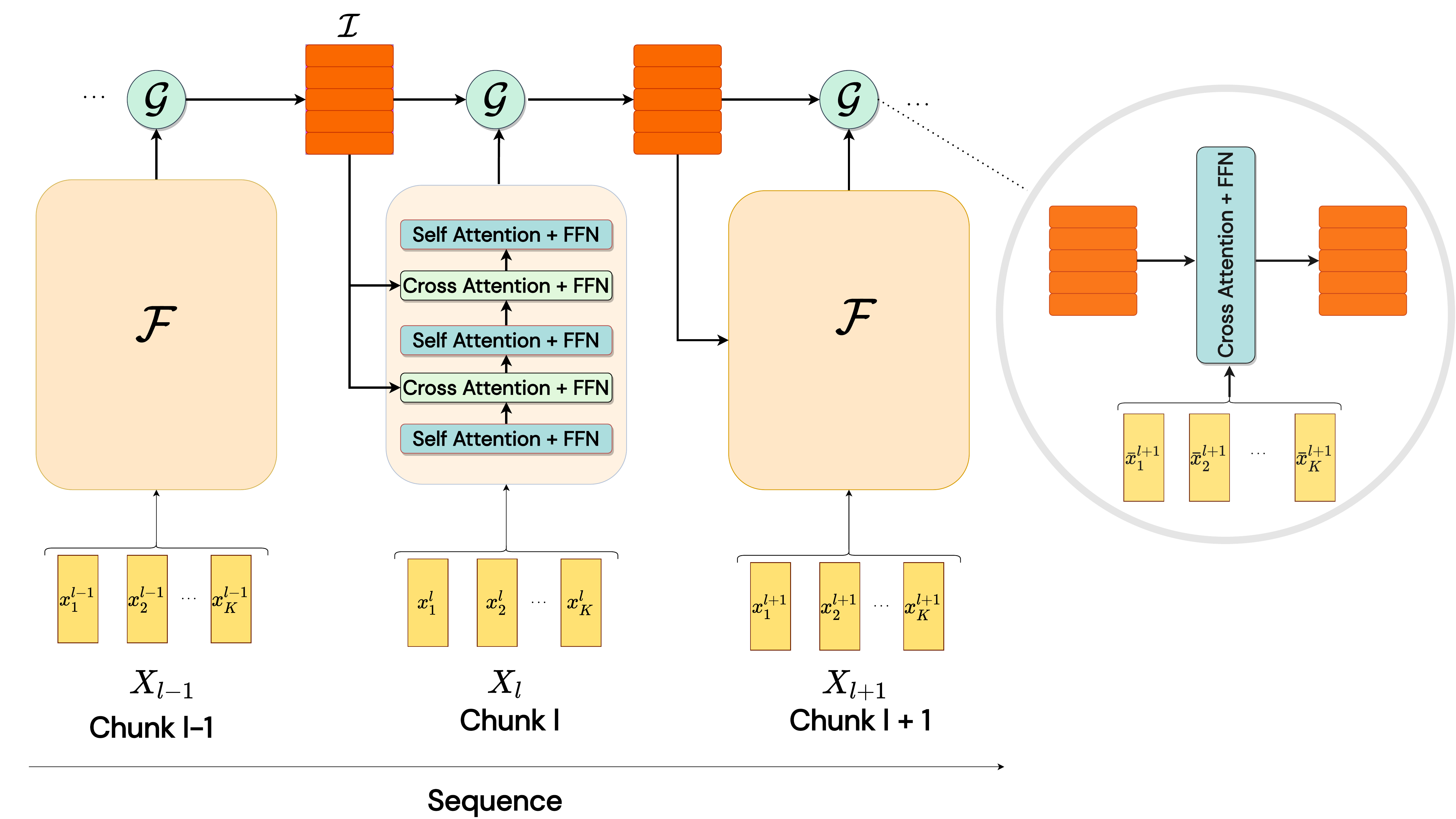}
  \end{center}
  \caption{ \textbf{Perceptual module + Temporal Latent Bottleneck Model}. $\cF$ denotes the perceptual module or the fast stream which is a Transformer. $\cI$ represents the temporal latent bottleneck state (consisting of a set of vectors) that are updated using a recurrent function denoted by $\cG$.  The given sequence is first divided into chunks of size $K$ and each chunk $X_l$ is processed by $\cF$ which consists of interleaved \textsc{Self Attention + FFN} (denoted in blue) and \textsc{Cross Attention + FFN} (denoted in green) layers. The \textsc{Cross Attention + FFN} layers allow the representation of $\cF$ to be conditioned on top-down information from $\cI$. The representations of the temporal latent bottleneck state is updated using the outputs of $\cF$ by a recurrent function $\cG$, which consists of a \textsc{Cross Attention + FFN} layer as shown in the circle.   }
\label{fig:tb_full_model}
\end{figure}

Implementation-wise, we divide the input into fixed size chunks (Figure \ref{fig:tb_full_model}). The fast stream operates within each chunk while the slow stream consolidates and aggregates information across chunks updating itself once per chunk. This leads to \textit{information asymmetry} between fast and slow stream as the fast stream contains fine-grained local information while the slow stream contains coarse-grained distant information. Such kind of information asymmetry has shown to improve generalization and adaptation performance of learned policies in the context of RL \citep{goyal2019infobot, galashov2019information}. The fast and slow streams interact with each other though bottleneck of attention. The division of computation into a fast and slow stream eliminates the need for capturing all possible pairwise interactions and thus introducing selectivity in the attention mechanism resulting in a much lower computational complexity which is not quadratic in the input size.  We show that the limited capacity of the slow stream and consolidation of information by a recurrent neural network prevents the model from capturing unnecessary information not useful for the downstream task. We evaluate the proposed model in a number of domains showing that it consistently outperforms competent baselines showing improved generalization to scenarios not seen during training.

%Deep Learning has seen a variety of architectures for processing sequential data \citep{hochrieter1997long, schuster1997bidirection, cho2014gru}.  Sequence models which only rely on a short fixed context window may be seen as a form of fast-processing mechanisms, because there is no dependence on the longer-term history.  Early research in deep learning found that robustness is substantially improved by using a recurrent neural network to represent the entire history of the sequence, with direct evidence found in the domain of handwriting synthesis \citep{graves2013generating}.  These recurrent neural networks which force the entire hidden state to be compressed into a single hidden state became highly successful across speech recognition, natural language processing, and other sequence domains \citep{graves2013speech, graves2008offline, graves2005framewise, sutskever2014sequence}.  Recurrent neural networks also saw significant advances in architectural inductive biases, such as a bias towards making small and selective changes to the hidden state on each step which helped improve its generalization performance  \cite{DBLP:conf/iclr/KruegerMKPBKGBC17, DBLP:conf/iclr/HenaffWSBL17, goyal2019recurrent}.  

%This is especially true in the task of machine translation, where content from the entire source text is necessary for producing a complete translation \cite{cho2014nmt}.

\vspace{-5mm}
\section{Methodology}
\vspace{-2mm}
\iffalse
\begin{wrapfigure}{r}{0.5\textwidth}
\vspace{-8mm}
  \begin{center}
    \includegraphics[width=0.48\textwidth]{Figures/tb_overview.pdf}
  \end{center}
  
  \caption{Demonstration of the proposed fast and slow stream model. The slow stream $\mathcal{G}$ updates by reading information from the fast stream $\mathcal{F}$. The slow stream also provides contextual information to the fast stream through top-down feedback denoted by red arrows. }
\label{fig:fast_and_slow}
\end{wrapfigure}
\fi
We now present the proposed  approach in detail. Our model jointly leverages the strengths of Transformers \citep{vaswani2017attention} and recurrent neural networks \citep{cho2014gru, hochrieter1997long}. %We show that our approach retains the high expressive power of Transformers while also being data-efficient. 
%The fast stream operates on raw inputs and is instantiated using a Transformer. We denote this Transformer by $\mathcal{F}$. %The slow stream operates at a higher level and updates at a lower frequency than the fast stream. Any recurrent function can be used to instantiate the slow stream. %such as an LSTM \citep{hochrieter1997long} or a universal Transformer \citep{dehgani2018universal}. We denote this function by $\mathcal{G}$. %A high-level demonstration of the proposed fast and slow stream model is presented in Figure \ref{fig:fast_and_slow}.
\setlength{\textfloatsep}{0pt}
\begin{algorithm}[h]
\SetAlgoLined
    \PyComment{$\mathcal{C}$(query, key, value): \textsc{Cross Attention + FFN layer}} \\
    \PyComment{$\mathcal{S}$(query, key, value): \textsc{Self Attention + FFN layer}} \\
    \PyComment{L: Num. Layers} \\
    \PyComment{R: Num. $\mathcal{C}$ per $\mathcal{S}$} \\
    \PyComment{X: Input sequence of length T. shape: [B x T x D]} \\
    \PyComment{$\mathcal{I}$: The Temporal Bottleneck} \\
    \PyComment{K: Chunk Size} \\
    \PyCode{} \\
    \PyCode{X = torch.chunk(X, K, dim = 1)} \PyComment{List of length $\floor{T/K}$ with each element of size [B x K x D]} \\
    \PyCode{}{} \\
    \PyCode{for $X_c$ in X:} \\
    \Indp   % start indent
    \PyCode{for l in range(L):} \\
    \Indp
    %\PyCode{} \\
     \PyCode{$X_c$ = $\mathcal{S}^l$($X_c, \quad X_c, \quad X_c$)} \\
     %\PyCode{} \\
     \PyCode{if l \% R == 0:} \\
     \Indp
     \PyCode{$X_c$ = $\mathcal{C}^{\floor{L/l}}$($X_c, \quad \mathcal{I}, \quad \mathcal{I} $)} \\
     \Indm

    \Indm
    \PyCode{$\mathcal{I} = \mathcal{C}(\mathcal{I}, \quad X_c, \quad X_c)$} \\

\caption{PyTorch-style pseudocode for proposed model}
\label{algo:tb_algo}
\end{algorithm}
\vspace{-2mm}
\subsection{Desiderata for Fast and Slow Streams of Processing}
\vspace{-2mm}
We give the detailed description of the proposed model in the next section. Here, we give an overview of our architecture and discuss some of its key properties. Given an input sequence, it is first divided into chunks of size $K$. Each chunk is processed by perceptual module represented by a Transformer (denoted as $\mathcal{F}$). While processing each chunk, $\cF$ is also conditioned on information from the Temporal Latent Bottleneck module $\cG$. The slow stream is a recurrent stream which has its own state consisting of a set of $N$ vectors (or slots) also called temporal latent bottleneck state denoted as $\cI$ in Figure \ref{fig:tb_full_model}. In the following sections, we use the term \textit{temporal latent bottleneck} to refer to the temporal latent bottleneck state $\cI$. This state is updated once per chunk using information from the perceptual module through a cross attention mechanism. 

%Formally, we call the slow stream the \textit{temporal latent bottleneck module} since it bottlenecks information from the past to the future. Being a recurrent stream, the slow stream has a state that updates over time. We implement the state of the temporal latent bottleneck module as a set of $N$ vectors called \textit{slots} and update it using using the information from the fast stream using a cross-attention mechanism. The fast stream is referred to as the \textit{perceptual module} since it processes the raw input (For eg. image, text, numbers etc.). We use a Transformer for the fast-stream and use an attention mechanism as the interface between the fast and slow streams.

%Having multiple streams of processing allows the model to represent information at \textbf{multiple scales}, which makes it easier to represent hierarchical information. Previous works have shown that temporal data has a natural hierarchical structure \citep{chung2016hierarchical, hihi1995hierarchical, mozer1991induction, Schmidhuber91neuralsequence}. The slow stream captures high level and slowly changing information while the fast steam captures low level and frequently changing information. For example, if text is input into the model as bytes, the fast stream will capture information at the scale of bytes while the slow stream will learn to capture character-level or world-level information. While most past work introduces hierarchies in recurrent neural networks, 

The perceptual module operates within each chunk while the temporal latent bottleneck operates across chunks slowly updating itself after each chunk has been processed by the perceptual module. Thus, the only way the perceptual module gets information about inputs beyond its own chunk is through the temporal latent bottleneck. An added advantage of this is that the computational complexity of the attention mechanism in the proposed model is $\cO(\frac{T}{K} (K^2 + KN))$ while that of a Transformer is $\cO(T^2)$, where $T$ is the length of the sequence, $K$ is the chunk size, and $N$ is the number of temporal latent bottleneck state vectors. Since $K << T$ and $N << T$, we can see that $\frac{T}{K} (K^2 + KN) < T^2$. Therefore the proposed model has a much lower computational complexity compared to a Transformer. Furthermore, the capacity of the temporal latent bottleneck is  limited and much smaller than that of the perceptual module. This encourages the temporal latent bottleneck to represent the most salient information about the past while the perceptual module represents only local information. This creates an \textit{information asymmetry} between the two streams. This information asymmetry leads to the perceptual module
having a fine grained view of the nearby inputs but a very coarse grained view of the distant past. This is very different from the usual self-attention which attends to all tokens in the sequence at the same level of granularity. 

An advantage of having a compressed representation of the past is that it allows the model to forget irrelevant information. For example, if an agent is navigating in a large maze, it does not need to have fine grained knowledge of its actions from the distant past. In the case of a Transformer, it would attend to every step from the past (including steps from the distant past) which may be irrelevant in the present context thus wasting its capacity in modeling irrelevant details.  Another important component of the proposed model is top-down attention which conveys contextual information from the high-level Temporal Latent Bottleneck module to the processing of low-level perceptual module. Past works \citep{mittal2020learning, fan2021addressing, hill2020grounded, dai2019transformer} have shown that top-down attention improves generalization and adaptation performance of the learned model. One difference between these works and the proposed model is that in their case the multiple streams operate at the same temporal granularity while in our case the streams operate at a different time scales (because of information asymmetry). Through our experiments, we show the advantage of the proposed architecture over these works. Next, we describe the detailed implementation of the proposed model.

\vspace{-2mm}
\subsection{Computational Steps}
\vspace{-2mm}
\label{sec:computational_steps}
We denote the input $X$ as a sequence of $T$ tokens - $X = [x_0, x_1, x_2, \ldots, x_t]$. We chunk this input into chunks of size $K$ resulting in $\lfloor T / K \rfloor$ chunks. We refer to $l^{th}$ chunk as $X_l$. We represent the state of the temporal latent bottleneck $\mathcal{I}$ (i.e. the slow stream) as a set of $M$ $d$-dimensional vectors. As mentioned previously, we denote the temporal latent bottleneck module as $\cG$ and the perceptual module as $\cF$. $\cG$ updates the temporal latent bottleneck state while $\cF$ processes chunks $X_l$ to form the latent representation $\bar{X}_l$ -

\vspace{-5mm}
\begin{align}
    \text{\textbf{Perceptual Module}} \quad \bar{X}_l &= \cF(X_l, \cI_l) \\
    \text{\textbf{Temporal Latent Bottleneck Module}} \quad \cI_{l + 1} &= \cG(\cI_l, \bar{X}_l)
\end{align}
%\vspace{-5mm}
%Next, we go into details of the implementation of these modules. 

\textbf{Preliminaries}. The central components of our model are the key value attention mechanism \citep{Bahdanau2015NeuralMT, vaswani2017attention} and the FFN module \citep{vaswani2017attention}. We use two forms of the attention mechanism -(1) Self Attention \citep{vaswani2017attention}: In this the query and key vectors refer to the same set of vectors; (2) Cross Attention \citep{goyal2021coordination, jaegle2021perceiver, goyal2019recurrent}: In this the query and key vectors refer to seperate sets of vectors.   %The attention mechanism allows to dynamically select information from a set of $\cN$ \textit{read} vectors $\cR$ to update a set of $\cM$ \textit{write} vectors $\cW$. The write vectors are first projected to a set of queries $Q = W^q\cW$, where $Q \in \mathbb{R}^{\cM \times D}$. The read vectors are projected into keys $K = W^k\cR$ and values $V = W^v\cR$, where $K \in \mathbb{R}^{\cN \times D}$ and $V \in \mathbb{R}^{\cN \times D}$. The updated write vectors $\bar{\cW}$ are obtained as - 

\textbf{Perceptual Module $\cF$.} As mentioned previously, the perceptual module refers to the fast stream that acts directly on the input. The perceptual module operates on each chunk separately. Therefore, at any time the input to the perceptual module are the tokens corresponding to a particular chunk $X_l = [x_{l \times K}, x_{l \times K + 1}, \ldots, x_{l \times K + K}]$. The perceptual module is a Transformer with self attention layers, cross attention layers, and FFNs. It has 2 kinds of layers - (1) \textsc{self attention + FFN}; (2) \textsc{cross attention + FFN}. The \textsc{self attention + FFN} layers process the input tokens and the \textsc{cross attention + FFN} layers integrate top-down information from the temporal latent bottleneck state $\cI$ as follows - 
\vspace{-5mm}

\begin{align} \label{eq:cross_transformer}
    X_l &= \textsc{Attention}(\text{LN}(X_l), \text{LN}(\cI), \text{LN}(\cI)) + X_l \nonumber \\
    X_l &= \textsc{FFN}(\text{LN}(X_l)) + X_l
\end{align}

We include one \textsc{cross attention + FFN} layer per $R$ \textsc{self attention + FFN} layers. The diagramatic representation of the perceptual module is presented in Figure \ref{fig:tb_full_model} (in the processing of chunk $X_l$). In the figure, we set $R = 1$.

%\begin{wrapfigure}{r}{0.5\textwidth}
%  \begin{center}
%    \includegraphics[width=0.44\textwidth, height = 4.5cm]{Figures/tb_inner_agent.pdf}
%  \end{center}
  
%  \caption{Demonstration of the temporal latent bottleneck module $\cG$. In this figure we use cross attention  + FFN modules for the temporal latent bottleneck. }
%\label{fig:inner_agent}
%\vspace{-5mm}
%\end{wrapfigure}

\textbf{Temporal Latent Bottleneck Module $\cG$.} The temporal latent bottleneck (TLB) module represents the slow stream that operates on the temporal latent bottleneck state $\cI$.  $\cI$ is updated using information from a particular chunk processed by the perceptual module. This update happens once for each chunk of the perceptual module resulting in $\lfloor T /  K \rfloor$ updates for $\cI$.  Since the temporal latent bottleneck state $\cI$ updates at a lower frequency than the perceptual module, it is expected to capture more stable and slowly changing features while the perceptual module captures faster changing features resulting in multiple scales of information representation. %As mentioned previously, the temporal latent bottleneck module also provides a bottlenecked view of the past for the perceptual module. 
An update to the temporal latent bottleneck state $\cI$ consists of a cross attention operation where the queries come from $\cI$ and the keys and values come from the output of the perceptual module. This cross attention operation is followed by an FFN update to $\cI$. Consider the perceptual module outputs for a chunk $l$ to be $\bar{X}_l = [\bar{x}_{l\times K}, \ldots, \bar{x}_{l \times K + K}]$. The update operation is implemented as follows:

\vspace{-5mm}
\begin{align}
     \bar{\cI} &= \textsc{Attention}(\text{LN}(\cI_l), \text{LN}(\bar{X}_l), \text{LN}(\bar{X}_l)) + \cI_l \nonumber \\
     \cI_{l + 1} &= \textsc{FFN}(\text{LN}(\bar{\cI})) + \bar{\cI}
\end{align}
\vspace{-6mm}

%In the case of a GRU style update, the output of cross attention serves as input to the GRU operation as follows: 
%\begin{align}
%     \bar{\cI} &= \textsc{Attention}(\text{LN}(\cI_l), \text{LN}(\bar{\vX}_l), \text{LN}(\bar{\vX}_l)) \nonumber \\
%     \cI_{l + 1} &= \textsc{GRU}(\bar{\cI}, \cI_l)
%\end{align}

The temporal latent bottleneck module introduces the notion of recurrence in our model. We show the details of this module in Figure \ref{fig:tb_full_model} (inside the circle). %We refer to the temporal latent bottleneck module as simply the \textit{temporal latent bottleneck} in the following sections of this paper.

\textbf{Perceptual Module + Temporal Latent Bottleneck Model.} We now present our complete architecture integrating both the perceptual module and the temporal latent bottleneck together. Given a sequence of tokens $X = [x_0, x_1, x_2, \ldots, x_t]$. We chunk this input into chunks of size $K$ resulting in $\lfloor T / K \rfloor$ chunks. The chunks are processed sequentially one after the other. For a chunk $k$, it is first processed using the perceptual module conditioned on information from the temporal latent bottleneck state. The outputs of the chunk are used to update the temporal latent bottleneck state $\cI$. The resultant temporal latent bottleneck state is then used to process the next chunk. The full model is presented in Figure \ref{fig:tb_full_model}. We use a Transformer as the perceptual module in our experiments. Thus our main contribution is introducing a temporal latent bottleneck into Transformers and showing its advantages through a variety of experiments. We also present the detailed algorithm for the proposed approach in Algorithm \ref{algo:tb_algo}.

%One drawback of the sequential nature of the proposed model is that a given token can only attend to future tokens in its own chunk and not beyond that while a Transformer can attend to all future tokens. This drawback does not affect the case where the given task is autoregressive in nature and only information from the past is available. For tasks where the entire future is available (for example - classification tasks) this may pose a problem. Through our experiments, we find that this drawback does not pose any major issues for the proposed model. We find that the proposed model convincingly outperforms Transformers that has access to the full future (and past).

%The final output from the model depends on task it is being used for. For a classification task, we take the mean across the vectors representing the final state of the temporal latent bottleneck and pass it to a classifier MLP which outputs the distribution over the labels. For an autoregressive task, where we have to output a label at each step, we use the output of the perceptual module $\bar{\vx}_l$ and pass it to an MLP which outputs the distribution over labels. 
\begin{wraptable}{r}{7cm}
    \vspace{-5mm}
    \centering
    \scriptsize 
    \setlength{\tabcolsep}{2pt}
        \caption{\textbf{Image Classification}. Here we compare the performance of the proposed  \textsc{ViT + TLB} model against \textsc{ViT} and \textsc{SwinV2} on CIFAR10 and CIFAR100 datasets for $64 \times 64$ images and $128 \times 128$ images. Note that the model is trained only on the $64 \times 64$ sized images and then transferred to $128 \times 128$ sized images. Results averaged across 3 seeds.}
        
    \label{tab:image_classification}
    \begin{sc}
    
    \begin{tabular}{|c|c|c|c|c|} 
\hline
& \multicolumn{2}{|c|}{\textbf{CIFAR10}} & \multicolumn{2}{|c|}{\textbf{CIFAR100}} \\
\hline
\textbf{Model} & $\bm{64 \times 64}$ & $\bm{128 \times 128}$ & $\bm{64 \times 64}$ & $\bm{128 \times 128}$ \\
\hline
\textsc{ViT} & 93.75 & 73.18 & 69.53 & 47.4 \\
%\hline
\textsc{Swin V2} & 97.66 & 84.9 & 79.95 & 58.59 \\
%\hline
\textsc{ViT + TLB} & 94.79 & 84.38 & 79.17 & 59.19 \\
\hline
\end{tabular}
    \end{sc}

    \vspace{-2mm}
\end{wraptable}
The proposed model is similar to a parallel work called Block Recurrent Transformers \citep{delesley2022block}. There are few differences between our work and theirs. First, they use a sliding window attention, while we divide the input into chunks. In their paper, they perform cross attention and self attention in parallel  while we  find that doing them sequentially and performing cross attention once per $R$ self attention steps yields better results. %They also use special tricks to deal with some instabilities in their case, while we find no such instabilities in our model. Also, while their focus is mainly on natural language tasks, we focus on a broader variety of tasks. 
We defer the rest of the discussion on related works to Appendix Section \ref{sec:related_work}

%The temporal latent bottleneck module in their case operates on one of the intermediate perceptual module layers while in our case the temporal latent bottleneck module operates on the output of the perceptual module. While they use an extremely high number of temporal latent bottleneck state slots (eg. 512), we find that using 5 or 10 slots yield good results. 

\vspace{-2mm}
\section{Experiments}
\vspace{-2mm}

%\begin{wraptable}{r}{7.0cm}
%\vspace{-10mm}
%\scriptsize    
%\caption{\textbf{Image Classification}. Here we compare the performance of the proposed  \textsc{ViT + TLB} model against \textsc{ViT} and \textsc{SwinV2} on CIFAR10 and CIFAR100 datasets for $64 \times 64$ images and $128 \times 128$ images. Note that the model is trained only on the $64 \times 64$ sized images and then transferred to $128 \times 128$ sized images. Results averaged across 3 seeds.}\label{tab:image_classification}
%\setlength{\tabcolsep}{3.3pt}
%\begin{sc}

%\begin{tabular}{|c|c|c|c|c|} 
%\hline
%& \multicolumn{2}{|c|}{\textbf{CIFAR10}} & \multicolumn{2}{|c|}{\textbf{CIFAR100}} \\
%\hline
%\textbf{Model} & $\bm{64 \times 64}$ & $\bm{128 \times 128}$ & $\bm{64 \times 64}$ & $\bm{128 \times 128}$ \\
%\hline
%\textsc{ViT} & 93.75 & 73.18 & 69.53 & 47.4 \\
%\hline
%\textsc{Swin V2} & 97.66 & 84.9 & 79.95 & 58.59 \\
%\hline
%\textsc{ViT + TLB} & 94.79 & 84.38 & 79.17 & 59.19 \\
%\hline
%\end{tabular}
%\end{sc}
%\vspace{-5mm}
%\end{wraptable}

%In this section, we outline the tasks in which we applied the temporal latent bottleneck and direct the reader to the appendix for more details on the experiments. 
\begin{wraptable}{r}{7cm}
    \vspace{-5mm}
    \centering
    \scriptsize 
    \setlength{\tabcolsep}{2pt}
        \caption{Here we show the performance of the proposed ViT + TLB model  against two baselines - One with no access to the past and One with no top-down information (i.e. high level to low level communication). We can see that the model suffers a drop in performance for both the baseline thus showing the importance of past information and top-down communication. Results averaged across 3 seeds.}
    \label{tab:ablation}
    \begin{sc}
    
\begin{tabular}{|c| c | c | c|c|} 
\hline
& & & \multicolumn{2}{|c|}{\textbf{CIFAR10}} \\
\hline
\textbf{Model} & Past  & Top &  $\bm{64 \times 64}$ & $\bm{128 \times 128}$ \\
& Info & Down & & \\
\hline

\textsc{ViT + TLB} & $\checkmark$ & $\checkmark$ & \highlight{94.79} & \highlight{84.38}\\
\textsc{No Past Info}. & $\times$ & $\times$  & 91.30  & 72.92\\
\textsc{No Top-Down Condn} & $\checkmark$ & $\times$ & 93.75 & 83.59 \\
%w/o BPTT & 17.19 & 13.28 \\
\hline

\end{tabular}
    \end{sc}

    \vspace{-2mm}
\end{wraptable}
Our goal is to show the wide applicability and benefits offered by the \textit{temporal latent bottleneck}, which we refer to as TLB. We demonstrate that the proposed model outperforms competitive baselines across many domains including vision, reinforcement learning, and natural language. Our main goal is to show that the proposed approach has high expressive power like Transformers while also being sample efficient unlike Transformers. Thus our main baselines are based on the original Transformer architecture. For example, we compare against ViT \citep{dosovitskiy2020vit} in image classification, Decision Transformer \citep{chen2021decision} in Reinforcement Learning, and Vanilla Transformer in rest of the tasks.  We also compare against some representative baseline that offer some of the key properties that our model offers. For example, we compare against state-of-the art Swin Transformer \citep{swinv2} which is a  strong baseline for image classification and is also hierarchical similar to the proposed model. We also compare against Transformer LS \citep{zhu2021long} which also processes long-term and short-term information using different attention streams. Furthermore, we also compare against Feedback Transformer \citep{fan2021addressing}, which also introduces top-down communication into Transformers.   Another key point of the proposed model is that any position cannot attend to any information from the future beyond its chunk since the temporal latent bottleneck only \textit{summarizes the past, not the future}. Meanwhile, \textbf{all the baselines we consider have bidirectional context} i.e. they can attend to all of the past and the future. We observe that despite this limitation, the proposed model outperforms all the considered baselines. %We first show that using the temporal latent bottleneck module in perceptual tasks like image classification leads to strong improvements and better generalization to high resolution images. We show that the proposed model is also an effective architecture for self-supervised learning. We also apply the temporal latent bottleneck in tasks that require the model to capture temporal dependencies across long time-scales and show that it outperforms strong baselines in such tasks. We then show the applicability of the proposed model in sequential decision making tasks including single-task and multi-task settings in setups such as Atari and BabyAI. \citep{maxime2018babyai}.

\vspace{-2mm}
\subsection{Temporal Latent Bottleneck For Perception}

\textbf{Image Classification.} Recently, Transformers have been widely applied for visual perception and have shown strong performance improvements over CNNs in tasks such as image classification, semantic segmentation, instance segmentation, etc. In this work we focus on image classification using Transformers. For a model to do well on image classification, it should learn to only focus on the relevant information and ignore other details (eg. background information). Self attention does not inherently have this inductive bias of ignoring irrelevant information since it models all pairwise interactions between the inputs. We posit that adding a limited bandwidth temporal latent bottleneck into the Transformer will allow the model to focus only on the most important information in the image which should enable the model to perform well. %Additionally, the top down feedback from the temporal latent bottleneck would also bias the perceptual module (a Transformer) to pay more attention to the most important parts of the image. We hypothesize that these properties of the temporal latent bottleneck module should allow it to perform well on image classification.

\begin{wrapfigure}{r}{0.4\textwidth}
\vspace{-2mm}
    \centering
\includegraphics[width=0.10\textwidth]{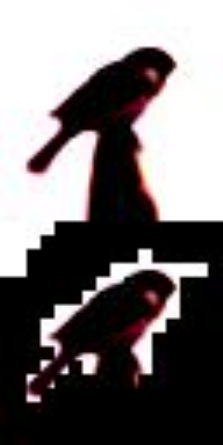}
\includegraphics[width = 0.10\textwidth]{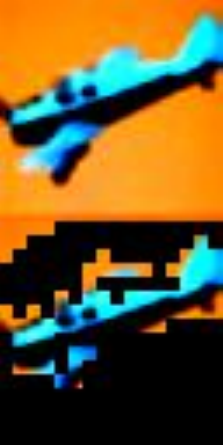}
    \caption{We want to analyze the information which is being used to influence the processing of Temporal Latent Bottleneck. To calculate this, we calculate the attention scores for different patches and we mask out all the patches that are not in the top 30\% of the attention scores. We can see that for both the images it recovers the foreground almost perfectly which shows it learns to focus on the most important information required to solve the downstream task.}
    \label{fig:image_map_vis}
    \vspace{-5mm}
\end{wrapfigure}

\textbf{Results}. We test our hypothesis on the CIFAR10 and CIFAR100 \citep{Krizhevsky09learningmultiple} image classification datasets. We also test the generalization abilities of the models by comparing their performance on images of higher resolution ($128 \times 128$) than seen during training ($64 \times 64$). We use ViT \citep{dosovitskiy2020vit} and Swin Transformer V2 (denoted as Swin V2) \cite{swinv2} as our baselines. Swin Transformer V2 has a key strength of generalizing to higher resolution images than those seen during training, making it a strong baseline.  The input image is split into patches of size $4 \times 4$ and fed in rastor order to all the models. For the proposed model we use ViT as the perceptual module and add a temporal latent bottleneck module to it. We call this model \textsc{ViT + TLB}. To predict the classification scores, we take the mean across the final temporal latent bottleneck state vectors and pass the resulting representation through an MLP.  We present the results for this experiment in table \ref{tab:image_classification}. \textit{We can see that \textsc{ViT + TLB} outperforms \textsc{ViT} for all cases and performs competitively to Swin Transformer V2}. For further hyperparameter details, we refer the reader to Appendix section \ref{appendix:image_classification}.  %Note that all the baselines can attend to information from all past and future tokens while ViT + TB can only attend to past tokens. Despite this limitation, the proposed model performs very well for image classification. 

\begin{wrapfigure}{r}{0.5\textwidth}
\vspace{-3mm}
    \centering
    \includegraphics[width=0.5\textwidth]{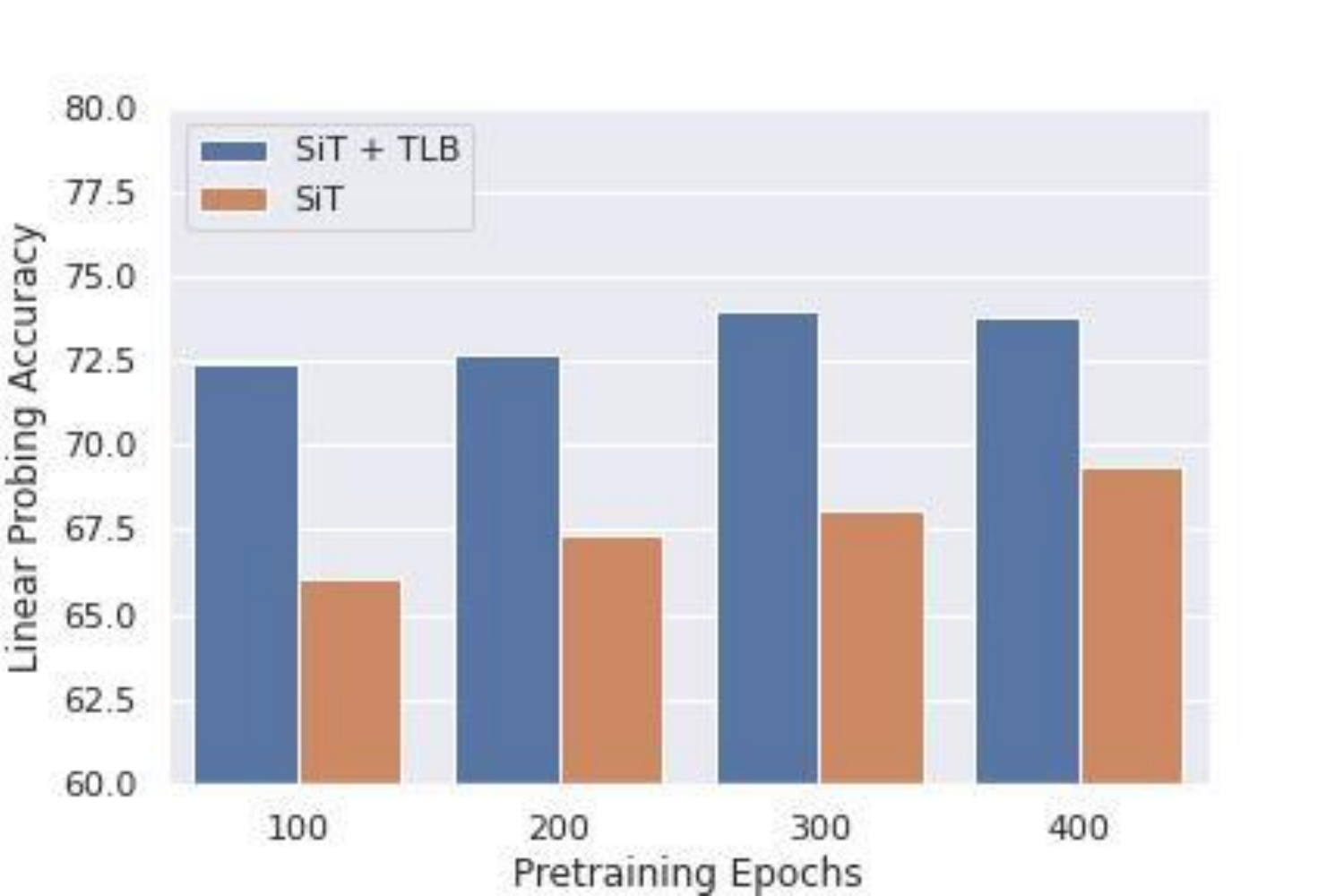}
    \caption{\textbf{Self Supervised Learning} Results of linear probing on the CIFAR 10 dataset for models pretrained on the STL 10 dataset. We can see that the proposed SiT + TLB approach outperforms SiT.}
    \label{fig:stl}
    \vspace{-7mm}
\end{wrapfigure}

\textbf{Quantitative Analysis}. One essential component of our model is top-down conditioning. Top down information helps in integrating information from the past as well as high-level information into the perceptual module. We hypothesize that both these kinds of information are important for the model to perform well. To test this, we design two baselines - (1) \textsc{ViT + TLB (No Past Info)}: In this baseline, we do not allow the TLB to communicate to the perceptual module, therefore the perceptual module has no information about the past; (2) \textsc{ViT + TLB (No Top-Down Condn)}: In this baseline, we have a separate temporal latent bottleneck module at every layer, therefore  the perceptual module has access to past information but does not have access to any high-level information through top-down feedback. We show the results for this ablation in Table \ref{tab:ablation}. We can see that the performance of both the baselines is worse than the proposed \textsc{ViT + TLB} model. This shows that both high-level information through top-down feedback and information from the past is important for the model to perform well.

\textbf{Qualitative Analysis}. To get a better understanding of what the temporal latent bottleneck is doing, we visualize the parts of the image where the temporal latent bottleneck pays most attention while it is being updated by the perceptual module. We present this visualization in Figure \ref{fig:image_map_vis}. We can see that the temporal latent bottleneck learns to pay the most attention to the foreground in each case. This further confirms our hypothesis that the limited capacity bottleneck focuses on the most important information required to solve the downstream task.

\textbf{Self Supervised Learning.} Many recent works have used Vision Transformers for self-supervised learning \citep{bao2021beit, ahmed2021sit, he2021masked, caron2021emerging, li2021mst, li2021efficient}. Here we show a proof-of-concept that introducing a temporal latent bottleneck in Vision Transformers results in better self-supervised representations. We consider the SiT model from \cite{ahmed2021sit} for this experiment. They use 3 objectives to pretrain their model - (1) The Reconstruction Objective - Reconstructs the input image, (2) The Rotation Prediction Objective - Predicts the rotation angle from [$0^{\circ}$, $90^{\circ}$, $180^{\circ}$, $270^{\circ}$], and (3) The Constrastive Objective (similar to SimCLR \citep{simclr}). For the proposed approach, we introduce a temporal latent bottleneck into SiT resulting in the SiT + TLB model. SiT also uses additional trainable contrastive and rotation tokens as input for calculating the contrastive and rotation objectives respectively. For SiT + TLB, we take the mean across the temporal latent bottleneck state vectors and use the resulting representation for computing the rotation and contrastive objectives. We use a chunk length of 20 for the SiT + TLB model. We pretrain the model for 400 epochs and evaluate the pretrained model at different epochs using linear probing.

\textbf{Results}. To evaluate the model, we pretrain the model on the STL10 dataset \citep{pmlr-v15-coates11a} and evaluate the learned representation using linear probing on the CIFAR10 dataset \citep{Krizhevsky09learningmultiple}. We present the results for this experiment in Figure \ref{fig:stl}. We can see that the proposed approach outperforms SiT thus showing the effectiveness of the proposed architecture for self-supervised learning. For additional experimental results and details, we refer the reader to Appendix section \ref{appendix:ssl}.

%We consider both the CIFAR10 dataset and the STL10 dataset for this setting. In the second setting we measure transfer across datasets. Here, we pretrain the model on STL10 dataset and then evaluate the learned representation using linear probing on the CIFAR10 dataset \citep{Krizhevsky09learningmultiple}

\vspace{-2mm}
\subsection{Temporal Latent Bottleneck for Sequential Decision Making}
\vspace{-2mm}

Transformers have recently been used for sequential decision making in reinforcement learning tasks such as Atari and BabyAI \citep{chen2021decision, iii2022improving}. These works deploy Transformers  in the offline RL setting where a large number of trajectories are available either through another trained agent or an expert agent. The Transformer is trained as an autoregressive generative model that predicts actions conditioned on the past context. We incorporate the temporal latent bottleneck module into the Transformer and explore its benefits in the RL setting. We test the proposed model in the  BabyAI \citep{babyai-env} and Atari \citep{atari-env} benchmarks. We describe our setups in detail below.

%We posit that the temporal latent bottleneck module will be useful in remembering the most relevant information that is useful for the current behavior of the agent and forgetting irrelevant details. For example, consider a large maze where the agent has to fetch a key to open a door. The key and the door maybe in different places in the maze. Once the agent has fetched the key and going towards the door, the path that it took to fetch the key becomes irrelevant. It only needs to the remember that the key has been fetched and it needs to start going towards the door. The temporal latent bottleneck provides a medium to forget such irrelevant information.

\textbf{Instruction Based Decision Making: BabyAI.} BabyAI \citep{babyai-env} provides a suite of environments where the agent has to carry out a given instruction in a partially-observable maze. These instructions include competencies such as going to an object in the maze, placing an object beside another object in the maze, opening a door with a key, etc. Some environments in the benchmark contain instructions that combine multiple competencies sequentially. For example, \textit{pick up a red ball and open the door in front of you after you pick up the grey ball on your left and pick up a red box}. Each environment in Baby AI benchmark has a different type of instruction that tests a different competency. The \textit{BossLevel} is the most complicated environment that contains instructions from all competencies. For more details regarding the various environments from the BabyAI benchmark, we refer the reader to Appendix section \ref{appendix:babyai}.

\begin{figure}[t]
    \vspace{-10mm}
    \includegraphics[width=0.9\textwidth]{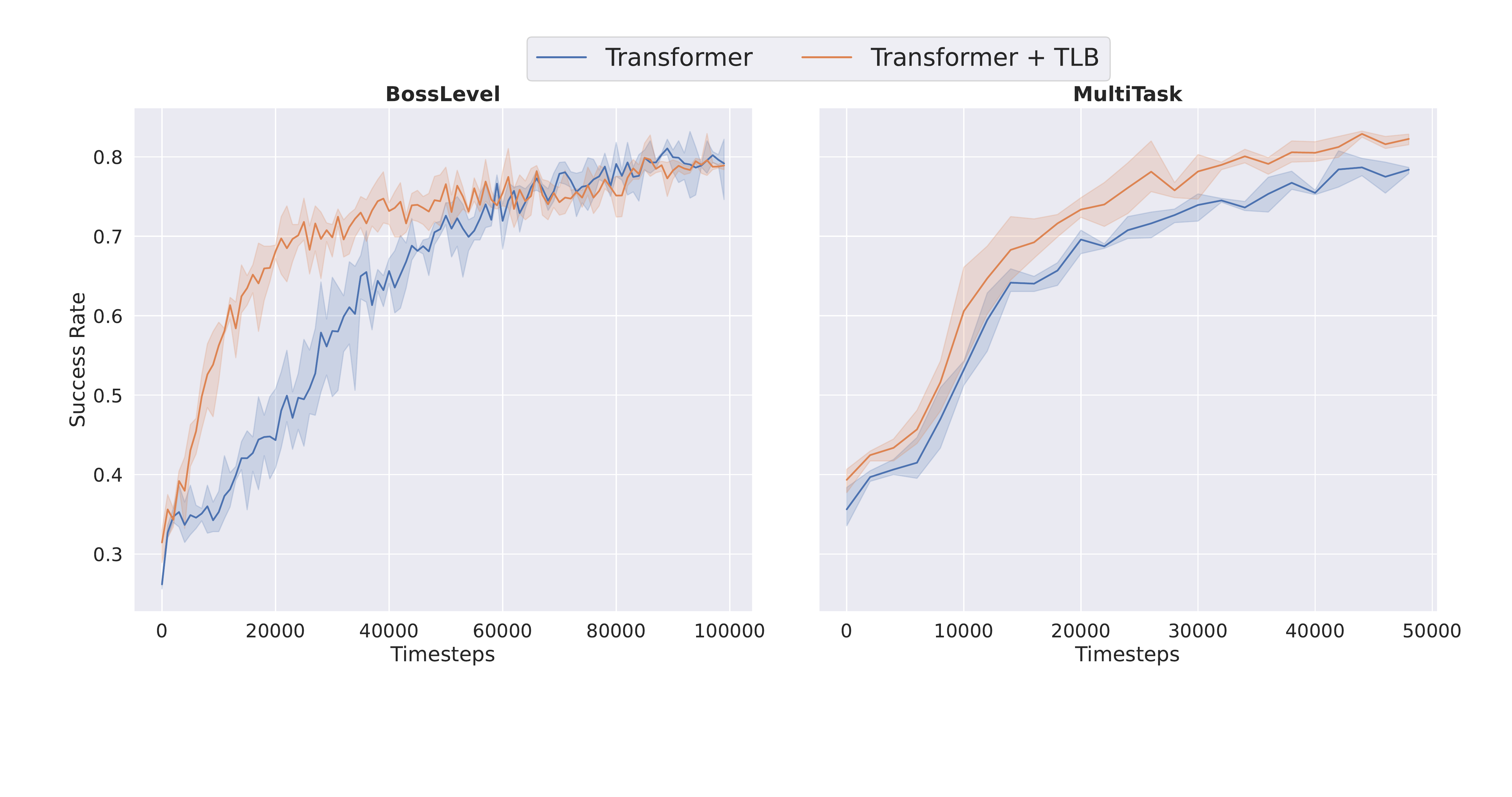}
    \vspace{-10mm}
    \caption{ \textbf{Single Task BabyAI}. (Left) Here we compare the performance of Transformer and Transformer + TLB on the BossLevel task from BabyAI. We can see that while both the models converge to a similar success rate, Transformer + TLB converges faster than Transformer. \textbf{Multi Task BabyAI}. (Right) Here we compare the performance of Transformer and Transformer + TLB on 8 tasks from the BabyAI suite of environments. A single model is trained for all the 8 tasks. We can see that Transformer + TLB converges faster and achieves a better performance than Transformer.}
    %CIFAR10 experiments on the integration of a fixed set of 100 specific examples for a single class that are repeated r times before moving on to the next class.}
    \label{fig:bosslevel}
\end{figure}

\iffalse
\begin{wrapfigure}{r}{0.4\textwidth}
%\vspace{2mm}
  \begin{center}
    \includegraphics[width=0.38\textwidth, trim  = {0 0 0 1.5cm}, clip]{Figures/babyai_multi_task_50000.pdf}
  \end{center}
  \caption{\textbf{Multi Task BabyAI}. Here we compare the performance of Transformer and Transformer + TLB on 8 tasks from the babyai suite of environments. A single model is trained for all tasks. We can see that Transformer + TLB converges and achieves a better performance than Transformer.}\label{fig:multi_task_50k}
  %\vspace{-5mm}
\end{wrapfigure}
\fi

We train our models with behavior cloning using expert trajectories from an oracle. For evaluation, we test the model by directly deploying it in the environment. We report the \textit{success rate} which measures whether the agent successfully carried out the given instruction or not. We use a Transformer \citep{vaswani2017attention} as the baseline in these experiments. For the proposed model, we introduce a temporal latent bottleneck into the Transformer-based perceptual module. For the baseline Transformer model, we append the language instruction to the sequence of states allowing the model to attend to the language instruction at each layer. For the proposed model, the language instruction is appended to each chunk, allowing each chunk to attend to it.

\iffalse
\begin{wrapfigure}{r}{0.4\textwidth}
\vspace{-5mm}
  \begin{center}
    \includegraphics[width=0.4\textwidth]{Figures/bosslevel.pdf}
  \end{center}
  \caption{\textbf{Single Task BabyAI}. Here we compare the performance of Transformer and Transformer + TLB on the BossLevel task from the BabyAI environment.}\label{fig:bosslevel}
 \vspace{-6mm}
\end{wrapfigure}
\fi

\textbf{Results}. We consider two settings - \textbf{Single task} and \textbf{Multi task}. In the single task setting, we evaluate the proposed approach on individual environments from the BabyAI benchmark while in the multi-task setting we train a single model on 8 different environments.

\textbf{Single Task.} We present the results for BossLevel in Figure   \ref{fig:bosslevel} (left) and present the results for the other tasks in Appendix Figure \ref{fig:babyai_single_task}. \textit{We can see that while Transformer and Transformer + TLB achieve almost similar performance at convergence. However, Transformer + TLB is much more sample efficient, converging much faster}. We posit that the temporal latent bottleneck module prohibits the model from paying attention to unnecessary information which allows it to converge faster.  %For each environment, we use 100k expert episodes. The model is trained for 50k steps. For more details regarding the environments and the training setup, we refer the reader to the appendix. 

\textbf{Multi Task.} We present the results for the multi task setting in Figure \ref{fig:bosslevel} (right). We train the model on 8 environments - PutNext, Unlock, Synth,  GoToSeq, SynthLoc, GoToImpUnlock, BossLevel. We evaluate the model on the same 8 environments. We report the average success rate across 8 games. \textit{We can see that the Transformer + TLB model converges faster and also outperforms the Transformer}. We refer the reader to the appendix for more details regarding the model and training.

\begin{wraptable}{r}{7cm}
    \vspace{-5mm}
    \centering
    \scriptsize 
    \setlength{\tabcolsep}{2pt}
        \caption{\textbf{Atari}. Here we show that adding a temporal latent bottleneck into decision Transformer improves performance across various atari games. Results are averaged across 10 seeds.}
    \label{tab:atari}
    \begin{sc}
    
    \begin{tabular}{|c|c|c|}
    \hline
         \textbf{Game} & \textbf{DT} & \textbf{DT + TLB}  \\
         \hline
         Breakout & \g{71.51}{20.58} & \highlight{\g{87.63}{16.24}} \\
         Pong & \g{13.68}{2.00} & \highlight{\g{14.71}{1.78}} \\
         Qbert & \g{3268}{1773.07} & \highlight{\g{5019.75}{1647.13}} \\
         Seaquest & \g{1039.11}{122.90} & \highlight{\g{1248.22}{86.62}} \\
         \hline
    \end{tabular}
    \end{sc}

    \vspace{-2mm}
\end{wraptable}
\textbf{Atari.} \citep{chen2021decision} recently introduced the Decision Transformer (DT) which learns to play various games in the Atari benchmark from suboptimal trajectories of a learned agent. Decision Transformer models the offline RL problem as a conditional sequence modelling task. The model uses a causal mask and supervised training to match the actions in the offline dataset conditioned on the future expected returns and the past history. This is done by feeding into the model the states, actions, and the return-to-go $\hat{R}_c = \sum_{c'=c}^C r_c$, where $c$ denotes the timesteps. This results in the following trajectory representation: $\tau = \big( \hat{R}_1, s_1, a_1, \hat{R}_2, s_2, a_2, \hat{R}_3, s_3, a_3, \ldots \big)$, where $a_c$ denotes the actions and $s_c$ denotes the states.  At test time, the start state $s_1$ and desired return $\hat{R}_1$ is fed into the model and it autoregressively generates the rest of the trajectory. Experimental results show that DT can leverage the strong generalization capabilities of Transformers and achieve the desired returns in a wide variety of tasks in Atari and OpenAI Gym, outperforming previous approaches in offline RL.
\begin{wraptable}{r}{7cm}
\scriptsize 
    \centering
    \setlength{\tabcolsep}{4pt}
        \caption{\textbf{Long Range Dependencies}. Here we compare the performance of the proposed model against the recently proposed long-short Transformer model \citep{zhu2021long} and the vanilla Transformer model \citep{vaswani2017attention}. We can see that the proposed model outperforms both the baselines thus showing the superiority of the proposed model in modelling long-range and hierarchical dependencies. Results averaged across 5 seeds.}
    \label{tab:lra}
    
    \begin{sc}
    \begin{tabular}{| c|c | c |}
    \hline
    \textbf{Model} & \textbf{ListOps} & \textbf{Text}   \\
    & & \textbf{Classification} \\
    \hline
    Transformer & \g{37.64}{0.0001} & \g{64.0}{0.0001} \\
    Transformer LS &\g{37.5}{0.0002} & \g{65.5}{0.0003}
 \\
    %\hline
    Transformer + TLB & \highlight{\g{38.2}{0.0001}} & \highlight{\g{82.08}{0.44}} \\
    \hline
    \end{tabular}
    \end{sc}
\end{wraptable}

We use the same setup as used in \citep{chen2021decision} for our experiments. We set the context length to a fixed number $C$. During training, $C$ timesteps from an episode are sampled and fed into the model resulting in a trajectory of length $3\vC$ (considering 3 modalities - returns-to-go, states, and actions). Each modality is processed into an embedding of size $d$. The state is processed using a convolutional encoder into an embedding of size $d$. The resulting trajectory is fed into the decision Transformer. The outputs corresponding to the states $s_c$ are fed into a linear layer to predict the action $a_c$ to be taken at timestep $c$. For the proposed model, we incorporate a temporal latent bottleneck module into the Decision Transformer. %The decision Transformer is used as the perceptual module. 

\textbf{Results}. We present our results in Table \ref{tab:atari}. The model is trained on 1\% of the Atari DQN-replay dataset \citep{agarwal2019striving} (500K transitions for each game). We use the same 4 games used in \citep{chen2021decision}: Pong, Seaquest, Qbert, and Breakout. \textit{We can see that the proposed model outperforms Decision Transformer in all the considered games thus showing the effectiveness of the proposed model}. More details regarding the model and training can be found in the appendix section \ref{appendix:atari}.
 
\vspace{-2mm}
\subsection{Temporal Latent Bottleneck for Long Range Dependencies}
\vspace{-2mm}

Here, we test the effectiveness of the proposed model in modelling long range dependencies. We apply the proposed model on the ListOps and text classification tasks from the Long Range Arena (LRA) benchmark \citep{yi2020long}. Both these tasks have very long sequences ranging from 1K to 4K tokens. Thus, for a model to do well, it has to learn to capture dependencies across very long time scales. Additionally, all these tasks have an inherent hierarchical structure. For example, Listops consists of nested list operations which makes it hierarchical. For text classification, the inputs consist of text in the form of bytes. Therefore, the model has to learn to compose bytes into characters and characters into words. We hypothesize that the multi-scale hierarchical nature of the proposed model will be extremely useful in modelling such hierarchical information. %The temporal latent bottleneck which operates at a slow scale can behave as a \textit{composer} that composes low-level information in relevant way to solve the desired task.

\begin{wrapfigure}{r}{0.6\textwidth}
    \centering
    \subfigure[ListOps]{
    \includegraphics[width=0.25\textwidth]{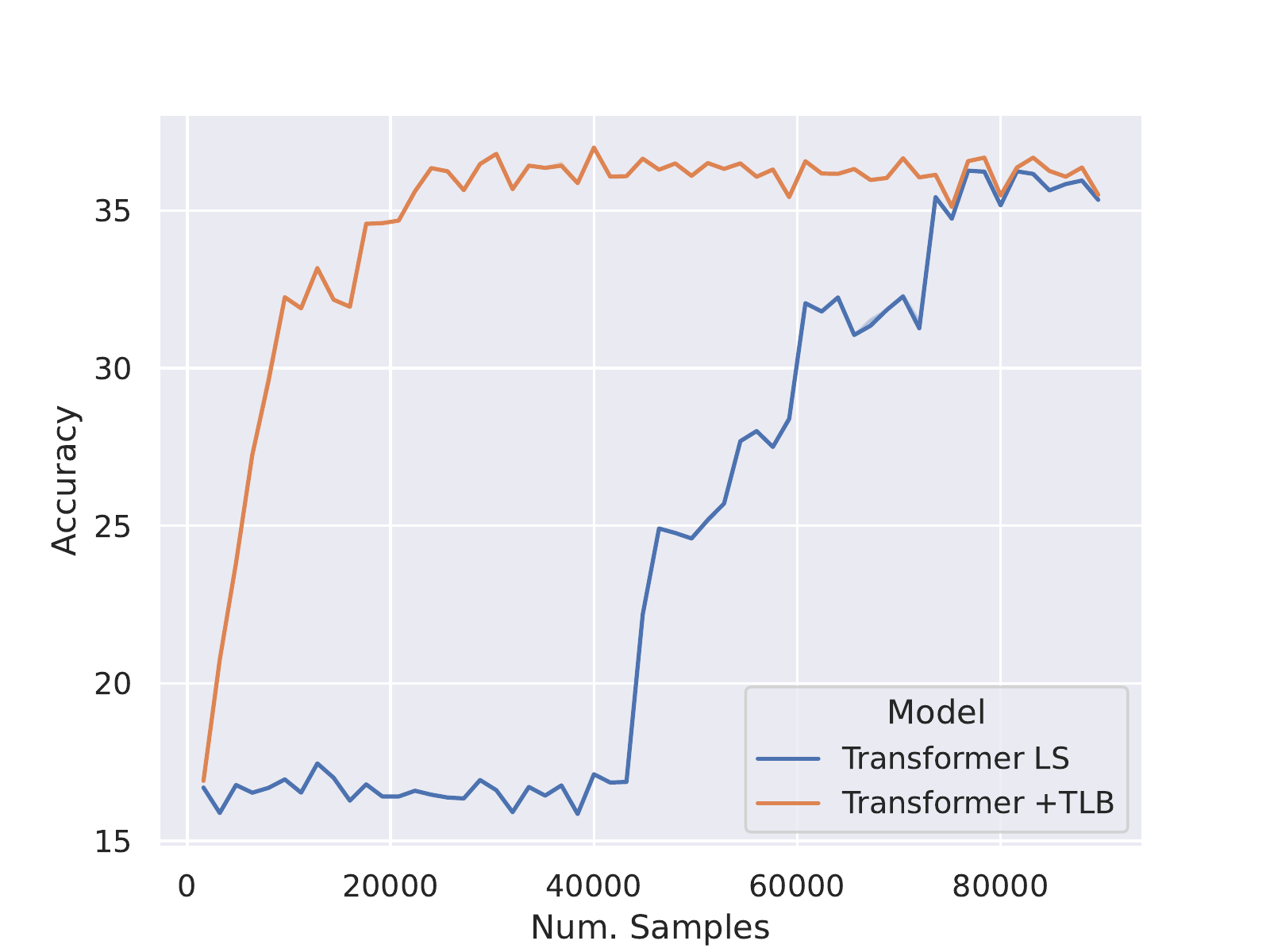}}
    \subfigure[Text Classification]{
    \includegraphics[width=0.25\textwidth]{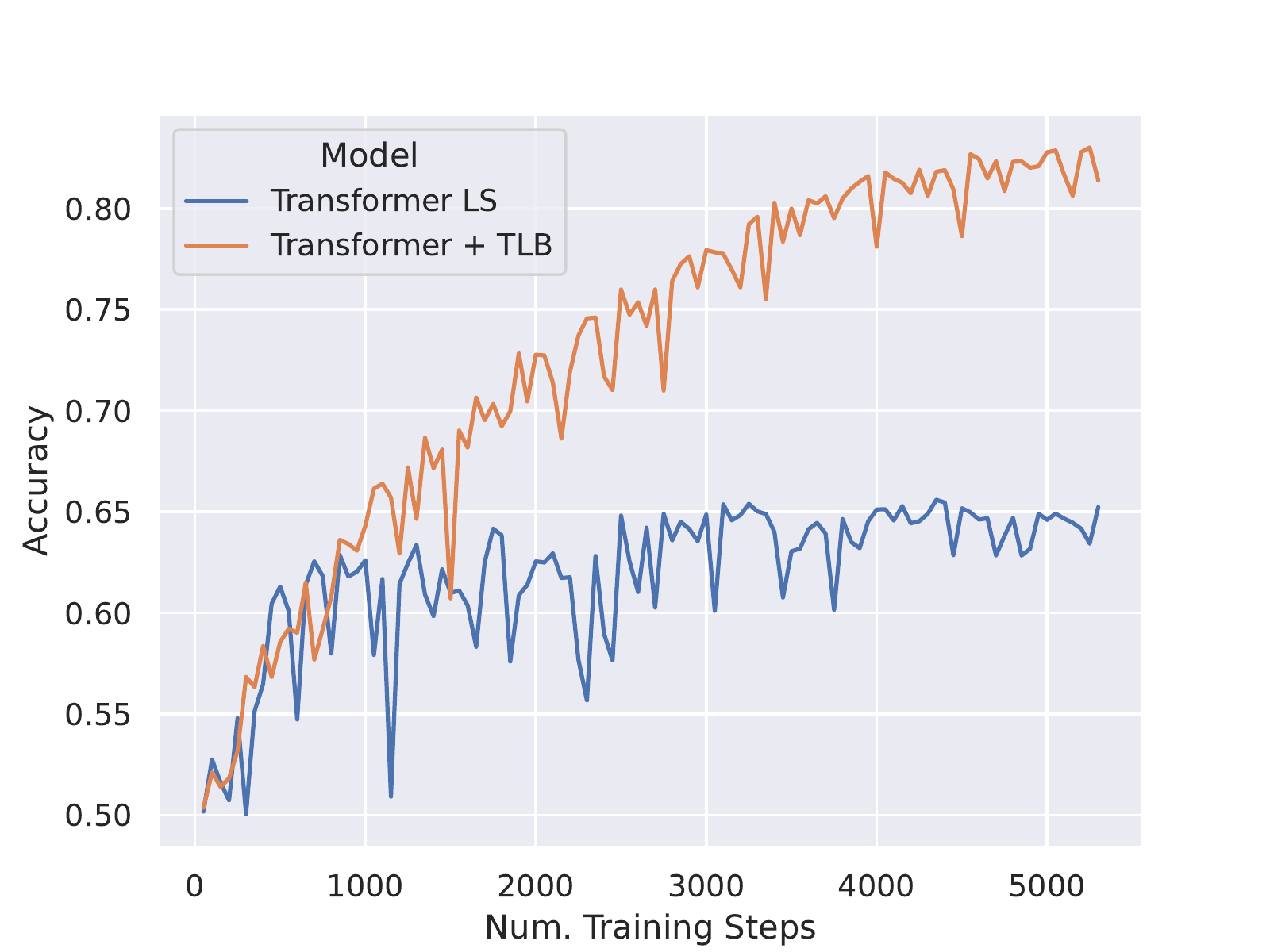}}

        \caption{ \textbf{(a)} Here we show the performance on ListOps as a function of the number of samples in the dataset. We do only one pass over the entire data and find that Transformer +TLB takes much fewer samples to converge as compared to the baseline Transformer LS. \textbf{(b)} Here we show the convergence curves of both the Transformer + TLB model and the Transformer LS model on the text classification task. In this case, we do not perform only one pass over the dataset since we observe that both models do not reach convergence in a single pass. Therefore, we report the number of training steps on the x-axis. We can see that the proposed model achieves much higher score than the baseline. }
    \label{fig:sample_efficiency_lra}
    \vspace{-3mm}
\end{wrapfigure}

\textbf{Results}. For this experiment, we use the same setup as% \textcolor{blue}{\cite{yi2020long} TODO###. Decide whether to cite yitay (text) or zhu (listops) here}  
\citep{zhu2021long}. For the proposed model, we use a Transformer as the perceptual model and implement the temporal latent bottleneck as described in Section \ref{sec:computational_steps}. We take the mean across the temporal latent bottleneck state vectors and use the resulting representation for classification. We compare the model against the long-short Transformer (\textsc{LS}) model \citep{zhu2021long}, which is a recently proposed model for the long range arena benchmark, and the vanilla Transformer model \citep{vaswani2017attention}. We present the results in Table \ref{tab:lra}. \textit{We can see that the proposed model outperforms both the baselines in both the tasks thus showing its usefulness in modeling long range dependencies}. %Furthermore, for text classification, we outperform the state-of-the-art model S4 model \citep{gu2022efficiently} which achieves a performance of 76.02\%.  
For further details, we refer the reader to Appendix section \ref{appendix:long_range_arena}.

\begin{wraptable}{r}{7cm}
\scriptsize 
    \centering
    \setlength{\tabcolsep}{4pt}
    \renewcommand{\arraystretch}{2}

        \caption{\textbf{Text Classification - Performance Ablation} Here, we compare the wall-clock time and memory during the training and inference phase of the text classification task w.r.t baseline transformer model.}
    \label{tab:text_efficiency}
    
    \begin{sc}
    \begin{tabular}{| c | c | c | c | c | c |}
    \hline
    \textbf{Chunk}  & 1000 & 100 & 40 & 20 & 10  \\
    \textbf{Size}  &  & & & & \\
    \hline
    Inference Speed &  3.5x & 3.6x & 3.3x & 2.2x & 1.2x\\
    Inference Memory &  0.09x & 0.08x & 0.12x & 0.08x & 0.1x\\
    Training Speed &  4.4x & 4.4x & 2.2x & 1.4x & 0.7x\\
    Training Memory &  0.14x & 0.08x & 0.49x & 0.40x & 0.42x \\

    \hline
    \end{tabular}
    \end{sc}

\end{wraptable}
In Fig. \ref{fig:sample_efficiency_lra}, we plot the convergence curves for ListOps and Text Classification. For ListOps (Figure \ref{fig:sample_efficiency_lra}(a)), we plot the convergence curves against the number of samples i.e. we do only one pass over the dataset hence the model does not see any example more than once. We can see that the proposed Transformer + TLB model is much more sample efficient than the baseline Transformer LS model. For Text Classification (Figure \ref{fig:sample_efficiency_lra}(b)), we plot the convergence curves against the number of training steps. We find that doing only one pass over the dataset does not work well for both the baseline and the proposed model hence we use number of training steps on the x-axis. We can see that while initially both models converge at a similar pace, the proposed model achieves a much higher performance.

We measure the wall-clock time and memory required for text classification task as we vary the chunk size in Table \ref{tab:text_efficiency}. All TLB models have an increased memory efficiency and supports faster inference speeds with respect to the baseline transformer model. The training speeds also get better with increased chunking. The only exception is very small chunk sizes, where the training is slower than the baseline because of increased temporal unrolling. However, as shown in Figure \ref{fig:sample_efficiency_lra}, such models are very sample efficient resulting in lesser training steps overall.

\textbf{Analysis}. Here we perform an ablation to show that the Temporal Latent Bottleneck does not only contain short-term information but also summarizes information from long term past. To test this hypothesis we design a baseline in which the current chunk attends to the previous few chunks instead of attending to the temporal latent bottleneck. We find that this baseline achieves a performance of \textbf{\g{32.10}{0.019}\%} compared to the proposed models \g{38.2}{0.0001}\% on the ListOps task. This shows that the Temporal Latent Bottleneck contains information about the long-term past. Additionally, here also we perform an experiment to probe the importance of top-down communication (i.e. high level to low level feedback). To do this we use the same Transformer  + TLB (No Top-Down Condn) baseline used in Table \ref{tab:ablation}. We find that this baseline achieves a performance of \textbf{\g{37.57}{0.003}\%} which is lower than the performance of the proposed Transformer  + TLB model which achieves \textbf{\g{38.2}{0.0001}\%} which further shows that top-down information from high-level to low-level is important for the model to perform well.

%\begin{wrapfigure}{r}{0.45\textwidth}
%\vspace{-6mm}
%  \begin{center}
%    \includegraphics[width=0.43\textwidth]{plots/copying.pdf}
%  \end{center}
%  \caption{\textbf{Copying Task}. Here we compare the performance of the proposed Transformer + TLB model to the Feedback Transformer model on the copying task. We can see that the Transformer + TLB achieves perfect accuracy for all the studied sequence length while the the performance of Feedback Transformer starts dropping after sequence length 400.}\label{fig:copying}
%  \vspace{-12mm}
%\end{wrapfigure}

We perform additional experiments to give us more insight into the behavior of the proposed model. We present these experiments in Appendix Section \ref{appendix:long_range_arena}. We also compare the model to additional efficient transformer baselines for all LRA tasks in Appendix Table \ref{tab:effecient_baslines_lra}.% In Tables \ref{tab:listops_chunk_ablation} and \ref{tab:listops_state_ablation} we analyse the effect of the chunk size and the number of state vectors on the performance of the model on the ListOps task. We find that for both cases, the model has an optimal number above or below which the performance drops. We explain our findings in more detail in Appendix Section \ref{appendix:long_range_arena}. We also plot the sample efficiency and convergence curves for ListOps and Text Classification respectively in Appendix Figure \ref{fig:sample_efficiency_lra}. We find that the proposed model is more sample efficient than the baseline Transformer LS for the ListOps task. %\textcolor{blue}{TODO Why not both the tasks here?}

\begin{figure}
\vspace{-12mm}
\centering
\begin{minipage}{0.45\textwidth}
\centering
\includegraphics[width=\textwidth]{plots/copying.pdf}
  \caption{\textbf{Copying Task}. Here we compare the performance of the proposed Transformer + TLB model to the Feedback Transformer model on the copying task. We can see that the Transformer + TLB achieves perfect accuracy for all the studied sequence lengths while the the performance of Feedback Transformer starts dropping after sequence length 400.}\label{fig:copying}

\end{minipage}
\hfill
\begin{minipage}{0.45\textwidth}
\centering
\captionsetup{type=table} %% tell latex to change to table
\setlength{\tabcolsep}{1pt}
\begin{tabular}{| c | c | c |}
    \hline
    \textbf{Sequence}  & \textbf{Feedback} & \textbf{Transformer + }   \\
    \textbf{Length} & \textbf{Transformer} & TLB\\
    \hline
    100 & 11800 & \highlight{6200} \\
    200 & 16600 & \highlight{9100} \\
    300 & 35100 & \highlight{12700} \\
   400 & NA & \highlight{14600} \\
  500 & NA & \highlight{13600} \\
   600 & NA & \highlight{19300} \\

    \hline
    \end{tabular}
      \caption{\textbf{Copying Sample Efficiency Ablation}. Here we present the number of unique samples required for the models to reach to perfect accuracy on the copying task. NA indicates that the model does not reach perfect accuracy. We can see that in all cases the Transformer + TLB model is more sample efficient than the Feedback Transformer model.}
    \label{tab:copying_sample_effeciecy}
\end{minipage}
  
\end{figure}

 \textbf{Temporal Latent Bottleneck for Copying Task.} Here, we study the copying task used in \citep{hochrieter1997long}. In the copying task, the model receives a sequence of 10 digits followed by blank inputs for a large number of steps, and then the model is asked to output the sequence of digits it received initially. Therefore, the model has to remember the original sequence of digits across long time scales. We can control the sequence length of this task by controlling the length of the blank input.% For example, for a copying task of sequence length 100, the model first receives a sequence of 10 digits between 1 and 8 followed by 100 zeros. The model then receives an indicator input which indicates that the model should start outputting the original sequence it received. The indicator in our case is the digit 9. After receiving the indicator, the model receives 10 more zeros and then it outputs the original sequence again.

The main motive behind studying this task is comparing the model to the Feedback Transformer model introduced in \citep{fan2021addressing} which also has top-down attention similar to the proposed model but does not represent information at multiple scales. We compare both the models on the copying task for sequence lengths 100, 200, 300, 400, 500, and 600. We present the results for this task in Figure \ref{fig:copying}. We can see that while both Transformer + TLB and Feedback Transform perform well for low sequence lengths, the performance of Feedback Transformer drops for longer sequence lengths above 400 while the proposed Transformer + TLB model still achieves perfect accuracy at long sequence lengths. We also compare the sample efficiencies to achieve perfect accuracy for both the models. We present this result in Table \ref{tab:copying_sample_effeciecy}. We can see that the proposed Transformer + TLB is more sample effecient than the baseline Feedback Transformer achieving perfect accuracy in much lesser number of samples in each case. For further details we refer the reader to Appendix Section \ref{sec:copying}.
 %The difference between the proposed model and Feedback Transformer is that the proposed model has two streams that represents information at different scales which creates an information assymetry in which the perceptual module represents fine-grained information from the nearby inputs while the Temporal Latent Bottleneck represents coarse-grained summarized information from the past. Meanwhile, Feedback Transformer processes information at a single scale only therefore it does not have any assymetry of information.

%\begin{wraptable}{r}{7cm}
%\vspace{-5mm}
%\scriptsize  
%    \centering
%    \setlength{\tabcolsep}{4pt}
%        \caption{\textbf{Copying Sample Efficiency Ablation}. Here we present the number of unique samples required for the models to reach to perfect accuracy on the copying task. NA indicates that the model does not reach perfect accuracy. We can see that in all cases the Transformer + TLB model is more sample effecient than the Feedback Transformer model.}
%    \label{tab:copying_sample_effeciecy}
    
%    \begin{sc}
%    \begin{tabular}{| c | c | c |}
%    \hline
%    \textbf{Sequence}  & \textbf{Feedback} & \textbf{Transformer + TLB}   \\
%    \textbf{Length} & \textbf{Transformer} & \\
%    \hline
%    100 & 11800 & \highlight{6200} \\
%    200 & 16600 & \highlight{9100} \\
%    300 & 35100 & \highlight{12700} \\
   % 400 & NA & \highlight{14600} \\
%    500 & NA & \highlight{13600} \\
 %   600 & NA & \highlight{19300} \\

%    \hline
%    \end{tabular}
%    \end{sc}
%\vspace{-5mm}
%\end{wraptable}

\section{Conclusion}
\vspace{-2mm}

We have developed an approach aimed at introducing selectivity in the interactions across time-steps in a transformer by splitting processing into two streams: (a) a slow stream that is updated in a recurrent manner and (b) a fast stream that processes the visual input.  The two streams are parameterized independently and interact with each other via attentional bottleneck. The information processed by the fast stream is used to change the state of the slow stream, and the information in the slow stream is used by the fast stream as contextual information to process the input. Through our experiments we show that the proposed approach works well across wide range of domains and problems. One limitation of the proposed model is that the chunk size is fixed and treated as a hyperparameter which requires some domain knowledge. Future work should explore methods for dynamic chunking.

\section{Acknowledgement}
The authors would like to thank Compute Canada for providing the computational resources used in this project. The authors also gratefully acknowledge the funding from Samsung, IBM and CIFAR.

%he proposed model aims at introducing selectivity in the interactions by dividing computation into two streams - (1) A high-level slow stream consisting of a set of slots updated in a recurrent manner (also referred to as the temporal latent bottleneck), and,  (2) A low-level attention-based fast stream that processes the input. The fast stream processing locally neighbouring information within chunks, while the slow stream contains information about distant tokens across chunks. The slow stream and the fast stream interact using a multi-head attention mechanism to achieve selectivity in how local and distant information is mixed.  We show that the resulting model substantially outperforms Transformers and shows improved generalization, especially when the test data includes novel challenges for the model that were not encountered during training.  

%\textbf{Ethics Statement}. The authors do not foresee any negative social impacts of this work, but of course the accumulation
%of improvements in ML could be misused as it may give more power to nefarious agents.
\clearpage

\bibliography{iclr2022_conference}
\bibliographystyle{iclr2022_conference}

\newpage
%%%%%%%%%%%%%%%%%%%%%%%%%%%%%%%%%%%%%%%%%%%%%%%%%%%%%%%%%%%%
\section*{Checklist}

%%% BEGIN INSTRUCTIONS %%%

Please do not modify the questions and only use the provided macros for your
answers.  Note that the Checklist section does not count towards the page
limit.  In your paper, please delete this instructions block and only keep the
Checklist section heading above along with the questions/answers below.
%%% END INSTRUCTIONS %%%

\begin{enumerate}

\item For all authors...
\begin{enumerate}
  \item Do the main claims made in the abstract and introduction accurately reflect the paper's contributions and scope?
    \answerYes{}
    \item Did you describe the limitations of your work?
    \answerYes{}
  \item Did you discuss any potential negative societal impacts of your work?
    \answerYes{}
    \item Have you read the ethics review guidelines and ensured that your paper conforms to them?
    \answerYes{}
\end{enumerate}

\item If you are including theoretical results...
\begin{enumerate}
  \item Did you state the full set of assumptions of all theoretical results?
    \answerNA{}
        \item Did you include complete proofs of all theoretical results?
    \answerNA{}
\end{enumerate}

\item If you ran experiments...
\begin{enumerate}
  \item Did you include the code, data, and instructions needed to reproduce the main experimental results (either in the supplemental material or as a URL)?
    \answerYes{}
  \item Did you specify all the training details (e.g., data splits, hyperparameters, how they were chosen)?
    \answerYes{Appendix Section \ref{sec:details}}
        \item Did you report error bars (e.g., with respect to the random seed after running experiments multiple times)?
    \answerYes{}
        \item Did you include the total amount of compute and the type of resources used (e.g., type of GPUs, internal cluster, or cloud provider)?
    \answerYes{}
\end{enumerate}

\item If you are using existing assets (e.g., code, data, models) or curating/releasing new assets...
\begin{enumerate}
  \item If your work uses existing assets, did you cite the creators?
    \answerYes{}
  \item Did you mention the license of the assets?
    \answerNo{All the datasets we use are openly available}
  \item Did you include any new assets either in the supplemental material or as a URL?
    \answerNo{}
  \item Did you discuss whether and how consent was obtained from people whose data you're using/curating?
    \answerNA{The datasets we use do not contain any sensitive information.}
  \item Did you discuss whether the data you are using/curating contains personally identifiable information or offensive content?
    \answerNA{}
\end{enumerate}

\item If you used crowdsourcing or conducted research with human subjects...
\begin{enumerate}
  \item Did you include the full text of instructions given to participants and screenshots, if applicable?
    \answerNA{}
  \item Did you describe any potential participant risks, with links to Institutional Review Board (IRB) approvals, if applicable?
    \answerNA{}
  \item Did you include the estimated hourly wage paid to participants and the total amount spent on participant compensation?
    \answerNA{}
\end{enumerate}

\end{enumerate}

%%%%%%%%%%%%%%%%%%%%%%%%%%%%%%%%%%%%%%%%%%%%%%%%%%%%%%%%%%%%

\section*{\Large Appendix}

\vspace{-2mm}
\section{Related Work}\label{sec:related_work}
\vspace{-2mm}

%Having multiple streams of processing allows the model to represent information at \textbf{multiple scales}, which makes it easier to represent hierarchical information. Previous works have shown that temporal data has a natural hierarchical structure \citep{chung2016hierarchical, hihi1995hierarchical, mozer1991induction, Schmidhuber91neuralsequence}. The slow stream captures high level and slowly changing information while the fast steam captures low level and frequently changing information. For example, if text is input into the model as bytes, the fast stream will capture information at the scale of bytes while the slow stream will learn to capture character-level or world-level information. While most past work introduces hierarchies in recurrent neural networks, 
\textbf{Hierarchical or Multiscale Recurrent neural networks.} This work takes inspiration from a wide array of work on introducing multiple scales of processing into recurrent neural networks \citep{chung2016hierarchical, hihi1995hierarchical, mozer1991induction, Schmidhuber91neuralsequence, jan2014clockwork}. These works divide the processing into multiple streams each operating at a different temporal granularity. While these works mainly focus on recurrent neural networks and their application is mainly on natural language tasks, we focus on introducing multiple streams of processing and a hierarchical structure into Transformers while also focusing on a broader range of domains beyond natural language. 

\textbf{Transformers.} Some of the components we describe in the proposed model have been used previously in various Transformer models. Transformer XL \citep{dai2019transformer} also divides the input into segments. Each segment considers the tokens from the current segment and the previous segment for attention without passing gradients into the previous segments. %This is loosely equivalent to the proposed model with a stop-gradient on the temporal latent bottleneck state after each chunk. %Another difference between the proposed model and Transformer XL is that the segment lengths they consider are much larger than the chunk lengths we consider. 
A number of previous works \citep{zhang2021multiscale, lu2021swin, wu2021cvt, yuan2021incorporating, wang2021pyramid, yang2021focal} have worked on introducing a hierarchical structure in Transformers mainly in the domain of vision. The main goal of these works has been to introduce convolution-like hierarchies into Vision Transformers \citep{dosovitskiy2020vit}. While these works progressively reduce the spatial resolution of the inputs in order to introduce hierarchies, we introduce hierarchies by adding another slow stream of information processing and without reducing the spatial resolution of the inputs. We also provision for the higher level of the hierarchy (i.e. the slow stream) to provide information to the lower levels as top-down conditioning which is not possible in any of the previous works. 

%This idea has been introduced recently in \citep{yang2021focal} and \citep{zhu2021long}. The main difference between these papers and our instantiation is that they mix both fine-grained and coarse-grained information in a single stream while we have separate streams for both kinds of information interacting through cross-attention mechanisms.

\textbf{Top-Down Conditioning.} Top-down information is information propagated from higher to lower levels of the network. It represents the models beliefs of the world and provides context for interpreting perceptual information. \cite{mittal2020learning} and \cite{fan2021addressing} have shown the advantages of top-down conditioning in recurrent neural networks and Transformers respectively. These works focus on different streams of processing operating at the same temporal granularity and the top-down conditioning is provided by higher level streams to the lower level streams. In our case, the top-down conditioning for the perceptual module is provided by the high-level slow stream which operates at a slower temporal granularity. This allows the perceptual model to be affected by much more long term high level information  as compared to just short-term high level information as in the case of \cite{mittal2020learning} and \cite{fan2021addressing}.

\section{Additional Experimental Details} \label{sec:details}
In this section, we cover the experimental details including the hyperparameter details and the the detailed task setups. In Section \ref{appendix:image_classification}, we cover details for the Image Classification experiment also presenting ablations for the CIFAR10 dataset. In Section \ref{appendix:ssl}, we describe details of the self-supervised learning experiment also presenting results on the STL10 dataset and showing that the learned representations from the SiT + TLB model transfer better to CIFAR 10 than the baseline SiT model. In Section \ref{appendix:long_range_arena}, we describe the experimental details of our experiments on the ListOps and Text Classification tasks from the Long Range Arena benchmark \citep{yi2020long}. In Section \ref{appendix:babyai}, we describe details of our BabyAI experiments and also present results for various BabyAI environments. We also show an ablation where we vary the chunk size and examine its effect on model performance. In Section \ref{appendix:atari}, we present details of our experiments on the 4 atari games. %In Section \ref{appendix:language_modelling}, we present details of our language modelling experiments. 

\subsection{Image Classification} \label{appendix:image_classification}

%\begin{wraptable}{r}{6cm}
%\vspace{-10mm}
%\scriptsize	
%\setlength{\tabcolsep}{4.5pt}
%\caption{ Here we show the performance of the proposed ViT + TLB model without top-down conditioning on the CIFAR10 dataset. We can see that the model suffers a significant drop in performance  without top-down conditioning thus showing the importance of top-down conditioning. Results averaged across 3 seeds.}\label{tab:ablation_2}
%\begin{sc}
%\begin{tabular}{|c|c|c|} 
%\hline
%& \multicolumn{2}{|c|}{\textbf{CIFAR10}} \\
%\hline
%\textbf{Model} &  $\bm{64 \times 64}$ & $\bm{128 \times 128}$ \\
%\hline

%\textsc{ViT + TLB} & \highlight{94.79} & \highlight{84.38}\\
%w/o Top-Down Condn.  & 91.30  & 72.92\\
%%w/o BPTT & 17.19 & 13.28 \\
%\hline

%\end{tabular}
%\end{sc}
%\vspace{-6mm}
%\end{wraptable}
We use the CIFAR10 and CIFAR100 datasets \citep{Krizhevsky09learningmultiple} for this task. We use a 9-layered model for the ViT baseline and a 12-layered model for the Swin Transformer Baseline. For the proposed model, we use a 6 layered model with $R$ set to 2 i.e. we apply one \textsc{Cross Attention + FFN} per 2 \textsc{Self Attention + FFN}. The proposed model uses 9,649,546 parameters while ViT uses 10,253,578 and Swin Transformer uses 10,438,264. We use the AdamW optimzer for training with learning rate 0.001 and weight decay of 5e-2. We use a batch size of 128.   
We train all models for 100 epochs. The input to the model is an image of size $64 \times 64$ which is divided into a sequence of patches each of size $4 \times 4$. For \textsc{ViT + TB}, the sequence is further divided into chunks of 16 patches. For the case with $128 \times 128$ sized images, we still divide the image into patches of $4 \times 4$ and interpolate the positional embeddings to adapt to the resulting longer sequence length as done in \cite{dosovitskiy2020vit} and \cite{swinv2}. For the proposed model, in $128 \times 128$ case we use a chunk size of 64. We use 1 v100 to train the model. The training time for the proposed is 24 hours. We use 5 temporal latent bottleneck state vectors for the proposed model.

\begin{figure}
    \centering
    \includegraphics[width = \linewidth]{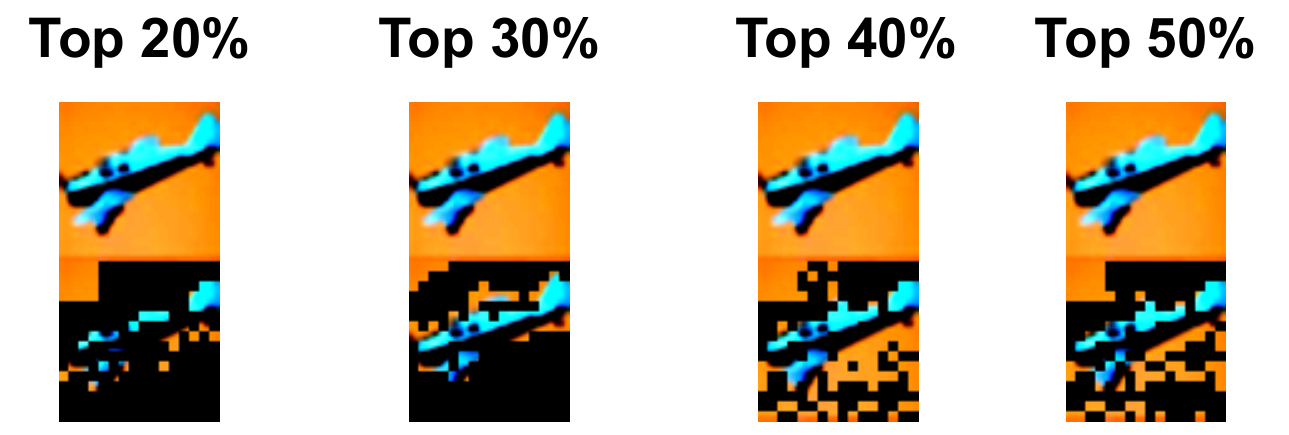}
    \caption{Here we visualize the patches that are in the top-k\% of the attention scores paid by the TLB. We can see that as we increase k, the TLB pays attention to a larger area of the image.}
    \label{fig:k_variation}
\end{figure}

We make a similar visualization to Figure \ref{fig:image_map_vis} in Figure \ref{fig:k_variation} but in this case we vary the attention threshold of the TLB. We mask out all the patches that are not in the top-k\% of the attention scores paid by the TLB. We can see that as we increase k, TLB pays attention to more patches which is expected.  

%\begin{figure}{r}
%\vspace{-6mm}
%  \begin{center}
%    \includegraphics[width=0.2\textwidth]{Figures/total269.pdf}
%    \includegraphics[width=0.2\textwidth]{Figures/total365.pdf} 
%    \includegraphics[width=0.2\textwidth]{Figures/total449.pdf}
%    \includegraphics[width=0.2\textwidth]{Figures/total527.pdf}
    
%  \end{center}
%  \caption{Here we visualize the parts of the image where the temporal latent bottleneck pays most attention to. We can see that it perfectly attends to the foreground in each case. This shows that the temporal latent bottleneck captures information from the most important parts of the input for the downstream task.}\label{fig:analysis}
%  \vspace{-5mm}
%\end{figure}
%In addition to Table \ref{tab:ablation}, we also test the proposed model without top-down conditioning on the CIFAR100 dataset. We report the results for this experiment in Table \ref{tab:ablation_2}. We can see that the performance of the proposed models drops without top-down conditioning thus underscoring its importance. 

%We analyse the information that is written into the temporal latent bottleneck. To conduct this analysis we mask out those parts of the image where the temporal latent bottleneck pays the the least attention. We present the results of this analysis in Figure \ref{fig:analysis}. We can see that the temporal latent background almost perfectly recovers the background in each case which shows that the it learns to most important features for the down stream task.

\subsection{Self Supervised Learning} \label{appendix:ssl}

For self supervised learning, we use the same setup as \citep{ahmed2021sit} and build on their codebase\footnote{https://github.com/Sara-Ahmed/SiT}. We use images of size $224 \times 224$. We use the same augmentation policy as \citep{ahmed2021sit}. For the baseline, we use a 12 layered Vision Transformer with 12 heads and embedding dimension 768. We use an FFN dimension of 3072. For the proposed model, we use a 12-layered model with $R = 2$. We use 5 temporal latent bottleneck state vectors. We use a patch size of 16 for both the models. For SiT + TLB, we use a chunk length of 20. Similar to \cite{ahmed2021sit}, we use the Adam optimizer with batch size 72, momentum 0.9, weight decay 0.05 and learning rate of 0.0005. We train the models for 5 days on 1 RTX8000 GPU completing 400 pretraining epochs. For models pretrained on the CIFAR10 dataset, we perform linear probing training for 500 epochs. For models pretrained on the STL10 dataset, we perform linear probing training for 300 epochs.

\begin{figure}%{r}{0.45\textwidth}
\vspace{-6mm}
  \begin{center}
    \subfigure[(a) STL10 Dataset]{\includegraphics[width=0.43\textwidth]{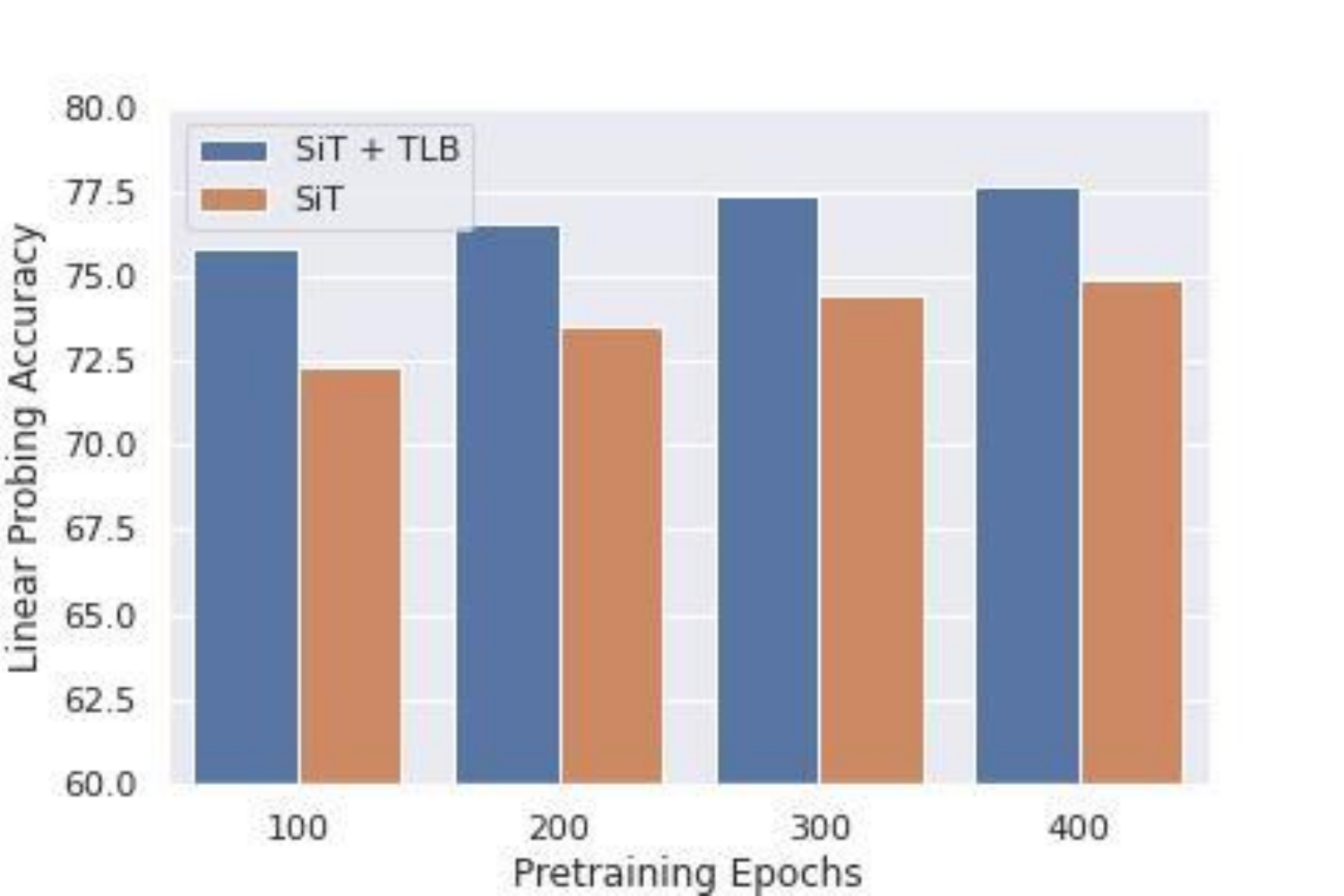}}
    \subfigure[(b) CIFAR10 Dataset]{\includegraphics[width=0.43\textwidth]{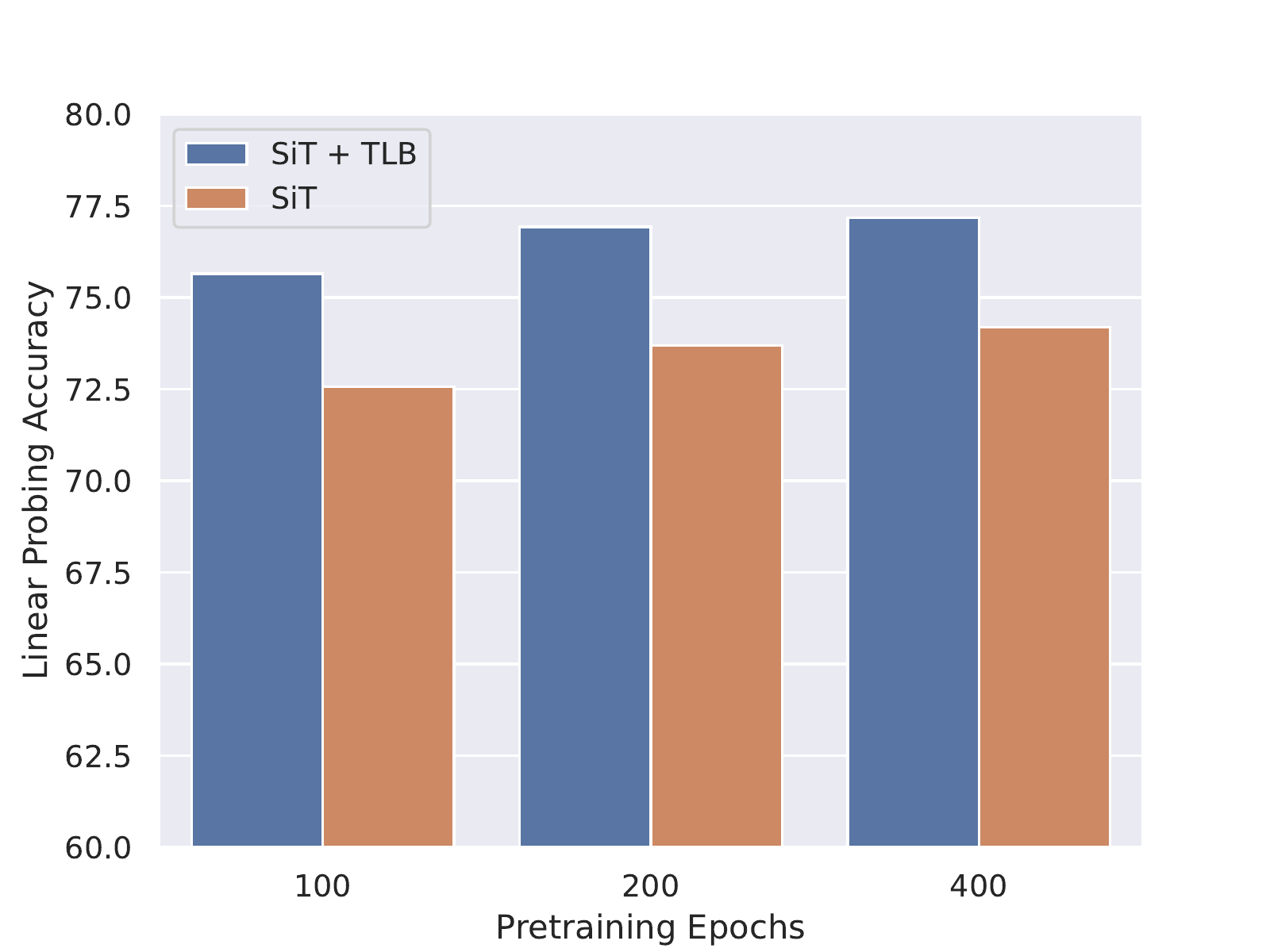}}
  \end{center}
  \caption{\textbf{Self Supervised Learning}. Here we show the result of training and linear probing on the STL10 and CIFAR10 dataset for different pre-training epochs. We can see that the proposed SiT + TLB approach outperforms SiT.}\label{fig:ssl}
\end{figure}

We present additional results for the self-supervised learning experiments in Figure \ref{fig:ssl}. In Figure \ref{fig:ssl}(a), we pretrain the model on the STL10 dataset and perform linear probing on the same dataset. In Figure \ref{fig:ssl}(b), we perform pretraining on the CIFAR10 dataset and perform linear probing also on CIFAR10. We can see that in both cases, the proposed SiT + TLB model outperforms SiT.

\begin{wraptable}{r}{7cm}
\scriptsize
\centering
\setlength{\tabcolsep}{1.0pt}
\renewcommand{\arraystretch}{2}
\caption{\textbf{ListOps - Chunk Size Ablation}. Here, we vary the chunk size in the listops task. We can see that there is optimal chunk size below and above which the performance drops. Results averaged across 5 seeds.}

    \begin{tabular}{|c|c|c|c|c|c|}
    \hline
         Chunk Size & 2  &  20 & 100 & 500 & 1000  \\
         \hline
        Acc & \g{36.08}{0.26} & \g{36.49}{0.16} & \g{38.18}{0.17} & \g{36.97}{0.27} & \g{36.82}{0.14}  \\ 
         \hline
    \end{tabular}
    
    \label{tab:listops_chunk_ablation}
\vspace{-5mm}

\end{wraptable}

\subsection{Long Range Dependencies} \label{appendix:long_range_arena}
For the experiments on the long range arena benchmark \cite{yi2020long}, we build on the codebase from \cite{zhu2021long} \footnote{https://github.com/NVIDIA/transformer-ls}. We describe our setups for the ListOps and text classification tasks below.

\paragraph{ListOps} We follow the same hyperparameters as \citep{zhu2021long}. In this task the model outputs the final number which is the result of a list operations. The number can be any number between 0-9, hence the model outputs a probability distribution over 10 possible numbers. For all the models, we use a transformer with embedding dimension 64 , FFN dimension 128 and 2 layers. For the Transformer + TLB model, we set $R = 1$. We use a chunk size of 20 and set the number of temporal latent bottleneck state vectors to 20. For training, we use Adam optimizer with a learning rate of 0.0001. We train the model for 5000 steps. We warmup the learning rate for 1000 steps. We use a batch size of 32.

In table \ref{tab:listops_chunk_ablation}, we run ablations on the chunk size of the proposed model. We can see that TLB shows maximum performance at chunk size = 100, and the performance gradually drops as chunk size reduces or increases. ListOps needs information from all of the input tokens and there is little redundancy in the data. Lower chunk sizes can potentially lead to a lot of information to integrate across chunks which might make the TLB forget important information more quickly and higher chunk sizes can lead to too much information to write in one chunk which can also lead to unwanted forgetting.

\iffalse
\begin{wraptable}{r}{7cm}
\vspace{-5mm}
\scriptsize	
    \centering
    \setlength{\tabcolsep}{4pt}
    \renewcommand{\arraystretch}{2}

        \caption{\textbf{ListOps - Chunk Ablation}. Here, we vary the chunk size in the listops task. Results averaged across 5 seeds.}
    \label{tab:listops_chunk_ablation}
    
    \begin{sc}
    \begin{tabular}{| c | c |}
    \hline
    \textbf{Chunk}  & \textbf{ListOps}   \\
    \textbf{Size} &  \\
    \hline
    2000 & \g{36.92}{0.27} \\
    1000 & \g{36.82}{0.14} \\
    500 & \g{36.97}{0.27} \\
    100 & \g{38.18}{0.17} \\
    20 & \g{36.49}{0.16} \\
    10 & \g{36.48}{0.36} \\
    5 & \g{36.24}{0.45} \\
    2 & \g{36.08}{0.26} \\
    \hline
    \end{tabular}
    \end{sc}
\vspace{-5mm}
\end{wraptable}
\fi

In table \ref{tab:listops_state_ablation}, we vary the number of state vectors and analyse its effect on the performance. We can see that there is an optimal number of state vectors (20) above or below which the performance drops. Less number of state vectors can lead to very low capacity in the Temporal Latent Bottleneck leading to loss of important information. Similarly, a high number of state vectors can lead to a high capacity in the Temporal Latent Bottleneck leading to it capturing a lot of unnecessary of noisy information.

The Temporal Latent Bottleneck is not only important since it provides information from the past but also since it provides high-level information through top-down feedback. To confirm this hypothesis we design an ablation study with two baselines. For the first baseline (baseline 1), we remove the Temporal Latent Bottleneck and only let the representations from the current chunk attend to the representations of the past chunk at the same layer (i.e. not high level to low level communication). For the second baseline (baseline 2), we introduce a temporal latent bottleneck at each layer. Each layer writes to its own TLB and reads from its TLB. Each layer-specific TLB provides summarized information from the past to the future chunks of that layer. This baseline is like the proposed model but without top-down communication i.e. the TLBs do not communicate any information to the lower levels. We find that baseline 1 achieves a performance of \g{32.10}{0.019}\% while baseline 2 achieves a performance of \g{37.57}{0.003}\%. The proposed model outperforms both the baselines achieving \g{38.2}{0.0001}\%. This shows that the top-down information provided by the Temporal Latent Bottleneck is an important component of the model and the TLB is not only a medium for providing information about the past.

\paragraph{Text Classification} Here we follow the hyperparameters as used in \citep{yi2020long}.  For all the models, we use a transformer with embedding dimension 256 , FFN dimension 1024. The baseline transformer has 4 layers with 4 transformer heads. For the Transformer + TLB model, we use 2 self-attention layers followed by 1 cross-attention for the fast step and 1 cross-attention layer for the slow step. We use a chunk size of 10 and set the number of temporal latent bottleneck state vectors to 10 unless otherwise specified. For training, we use the same learning rate schedule as used in \citep{yi2020long} but lower the learning rate by half to 0.025. We train the model for 20000 steps. We warmup the learning rate for 8000 steps. We use a batch size of 32. For text classification, we also add positional embeddings to the temporal latent bottleneck state vectors and local positional embeddings to each input chunk. We find that this is crucial for achieving good performance.

%\textcolor{blue}{##SHOULD WE REMOVE THIS LINE?##. The model is trained on TPU v2-8 architecture with a total memory of 64GB}
\begin{wraptable}{r}{7cm}
\scriptsize	
    \centering
    \setlength{\tabcolsep}{4pt}
    \renewcommand{\arraystretch}{2}

        \caption{\textbf{ListOps - Number of State Vectors Ablation}. Here, we vary the number of temporal latent bottleneck state vectors in the ListOps task. We can see that there is an optimal number of state vectors above or below which the performance drops. Results averaged across 5 seeds.}
    \label{tab:listops_state_ablation}
    
    \begin{sc}
    \begin{tabular}{| c | c | c | c | c |}
    \hline
    \textbf{State}  & 1 & 10 & 20 & 200   \\
    \textbf{Tokens} &  & & & \\
    \hline
    Acc &  \g{36.42}{0.32} & \g{37.16}{0.45} & \g{38.18}{0.17} & \g{37.25}{0.46}\\
  
    \hline
    \end{tabular}
    \end{sc}
\end{wraptable}

To further analyze the effect of chunking we perform an ablation where critical information is divided into two chunks. We introduce spaces in the input text such that each word is divided across two chunks. Therefore, for each chunk the perceptual module sees two half-words - the second half of the previous word and the first half of the next word. Note that the introduced spaces ensure that the chunk size is still 10. We find that the accuracy drops slightly from 82.08\% to 81.29\%. The very slight drop in accuracy shows that even when critical information is divided across chunks the model can still perform well. 

\paragraph{Comparison to Efficient Transformer Baselines} One of the claims of this work is that TLB has much lower computational complexity of attention than the vanilla transformer \citep{vaswani2017attention}. There has been a lot of work on reducing the computational complexity of the attention mechanism, especially for tasks involving long sequences such as the long range arena benchmark. We compare the proposed model against many efficient attention baselines on the tasks from the long-range-arena benchmark. We first give a brief description of all the tasks and then present the results in Table \ref{tab:effecient_baslines_lra}. 

\begin{itemize}
    \item \textbf{ListOps} - As mentioned previously, this task includes performing list operations on numbers such as Max and Min. This task contains sequences upto length 1024.
    \vspace{-1mm}
    \item \textbf{Text Classification} - As mentioned previously, this is a byte level text classification containing sequences of length upto 4k.
    \vspace{-1mm}
    \item \textbf{Retrieval} - This is also a byte-level text task. Here the model is tasked with outputting whether two documents are similar or not. The documents may be of lengths upto 4k.
    \vspace{-1mm}
    \item \textbf{Image Classification} - This is a pixel-level image classification task on the CIFAR10 dataset. Each image, when unrolled in a sequence, is of length 1024.
    \vspace{-1mm}
    \item \textbf{PathFinder} - This is also a pixel-level task containing images of size 32x32. When unrolled, the sequence length is 1024. The model is tasked with predicting whether two circles are reachable through a path containing dashed lines.
\end{itemize}

\begin{table}[]
\scriptsize
    \centering
    \begin{tabular}{|c|c|c|c|c|c|c|}
    \hline
    Model & ListOps & Text & Retrieval & Image & PathFinder & Avg \\
    \hline
     Linear Transformer \citep{linear_transformer} & 16.13     & 65.90 & 53.09     & 42.34     & 75.30     & 50.55 \\
     Reformer \citep{reformer}        &  \highlight{37.27}  & 56.10     & 53.40     & 38.07     & 68.50       & 50.67 \\
Sparse Transformer \citep{sparse-transformer}   & 17.07     & 63.58     &  59.59  &  44.24  & 71.71       & 51.24 \\
Sinkhorn Transformer \citep{sinkhorn-transformer} & 33.67     & 61.20     & 53.83     & 41.23     & 67.45       & 51.29 \\
Linformer \citep{linformer}       & 35.70     & 53.94     & 52.27     & 38.56     & 76.34       & 51.36 \\
Performer \citep{Performers}       & 18.01     & 65.40     & 53.82     & 42.77     &  77.05    & 51.41 \\
Synthesizer \citep{synthesizers}     & 36.99     & 61.68     & 54.67     & 41.61     & 69.45       & 52.88 \\
Longformer  \citep{longformer}    & 35.63     & 62.85     & 56.89     & 42.22     & 69.71       & 53.46 \\

BigBird  \cite{bigbird}       & 36.05     & 64.02     & 59.29     & 40.83     & 74.87       &  55.01  \\
\hline
Transformer + TLB & 37.05     &  \highlight{81.88}    &  \highlight{76.91}    &  \highlight{57.51}    & \highlight{79.06}    &  \highlight{66.48} \\
         \hline
    \end{tabular}
    \caption{In this table, we compare the performance of the proposed Transformer + TLB model to various transformer-based baselines that implement attention in an efficient manner. We can see that the proposed Transformer + TLB has the best overall performance by a significant amount. Results averaged across 5 seeds.}
    \label{tab:effecient_baslines_lra}
\end{table}

In Table \ref{tab:effecient_baslines_lra}, we can see that the proposed Transformer + TLB outperforms all efficient transformer baselines to achieve the best overall performance on the long range arena benchmark.

\begin{figure}
    \centering
    \includegraphics[width = \linewidth]{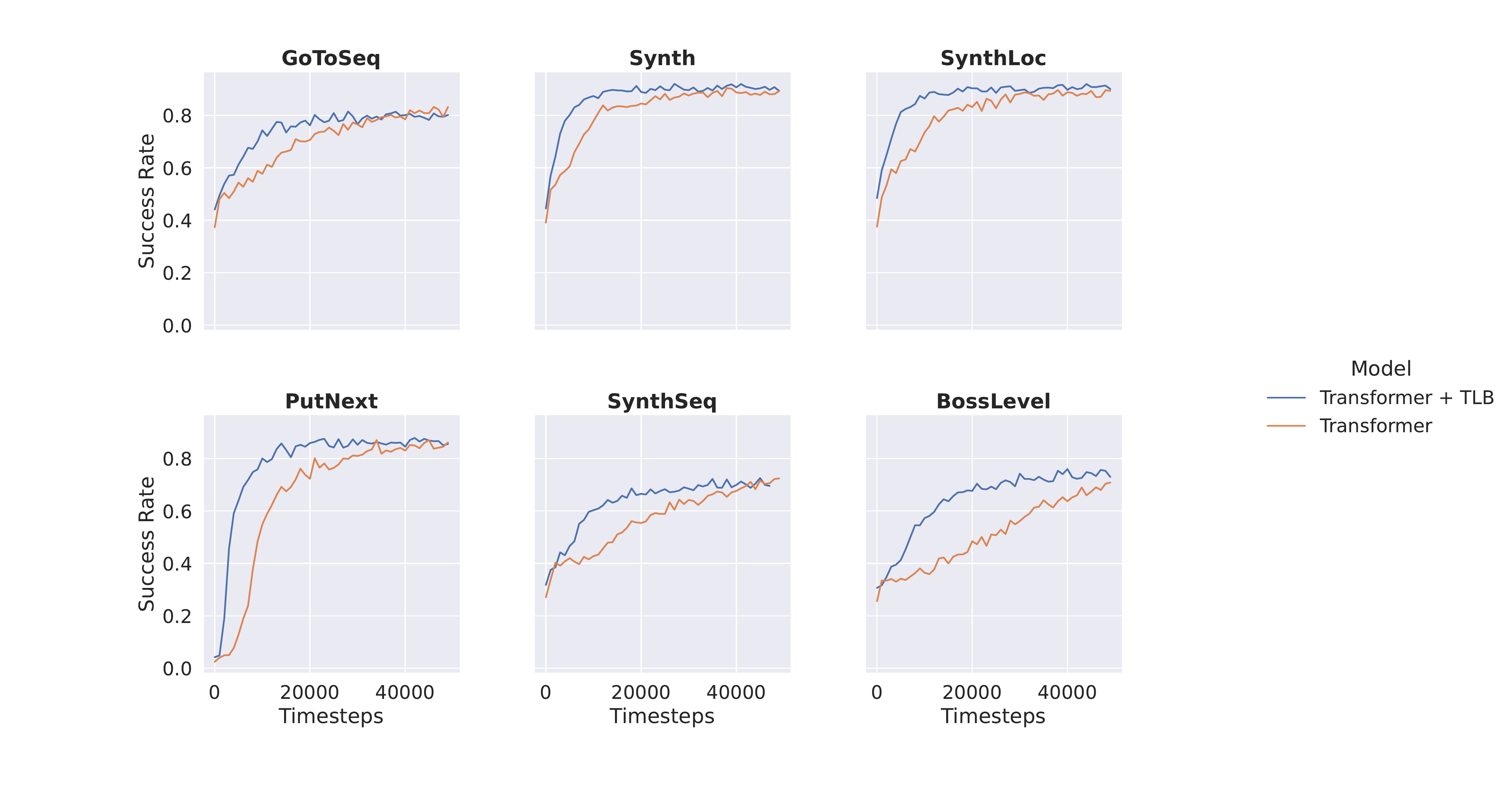}
    \caption{\textbf{Single Task Baby AI}. Here we compare the performance of Transformer and Transformer + TLB on individual tasks from the BabyAI benchmark. We can see that Transformer + TB converges much faster. Results averaged across 3 seeds.}
    \label{fig:babyai_single_task}
\end{figure}

\subsection{Baby AI} \label{appendix:babyai}

\begin{figure}
    \centering
    \includegraphics[width = 0.30\textwidth, trim = {2cm 0 2cm 0}, clip]{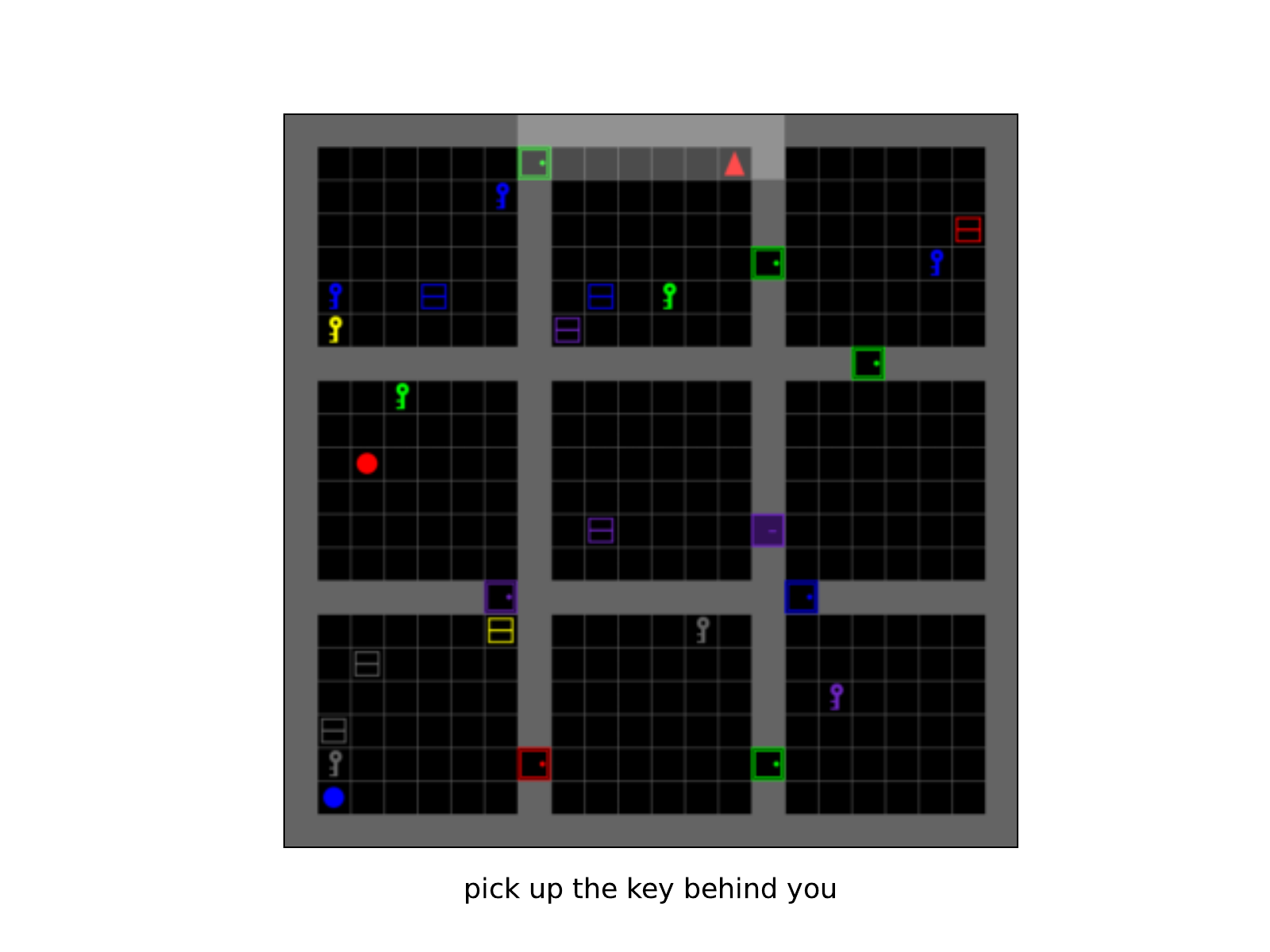}
    \includegraphics[width = 0.30\textwidth, trim = {2cm 0 2cm 0}, clip]{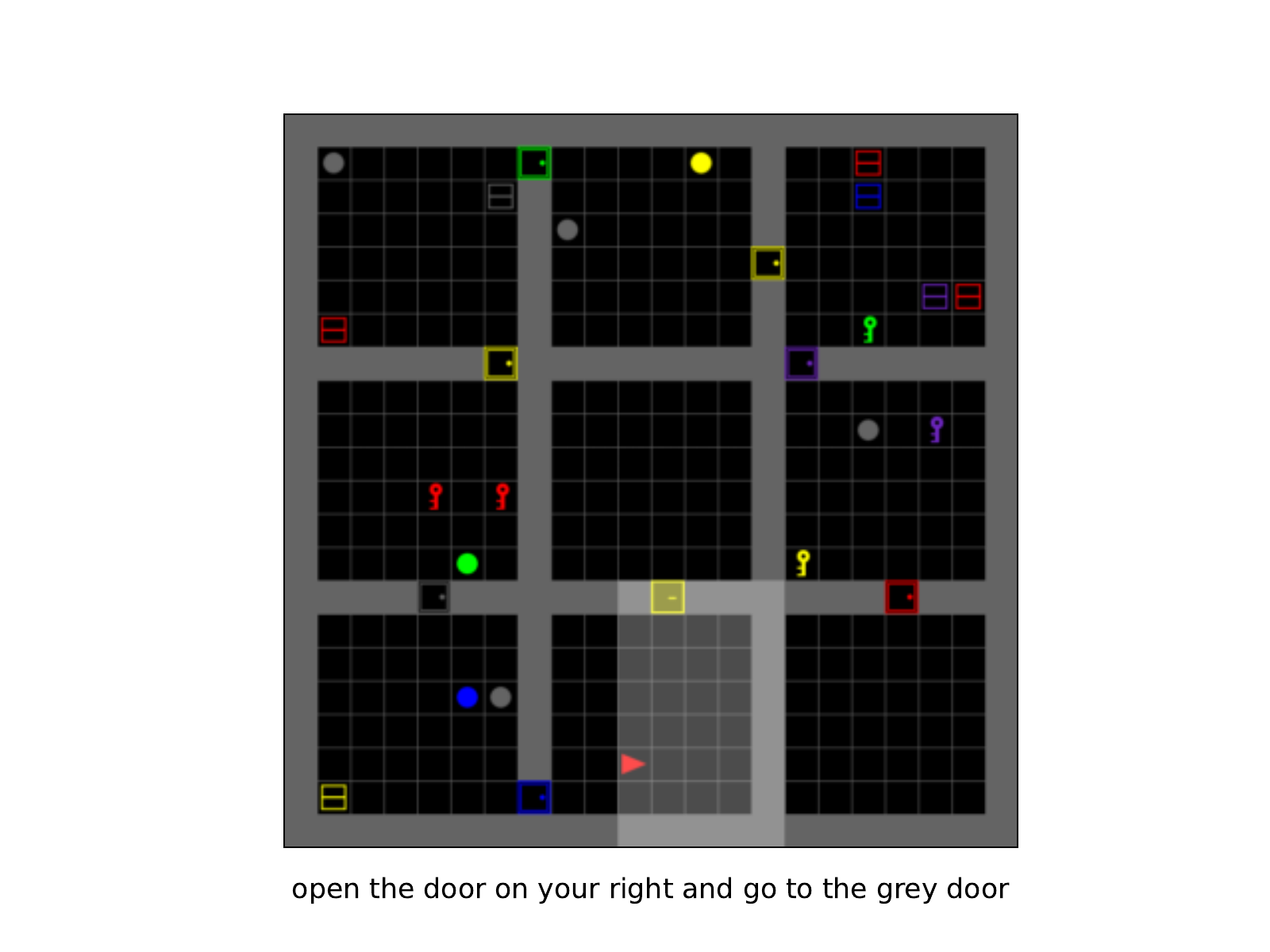}
   \includegraphics[width = 0.30\textwidth, trim = {2cm 0 2cm 0}, clip]{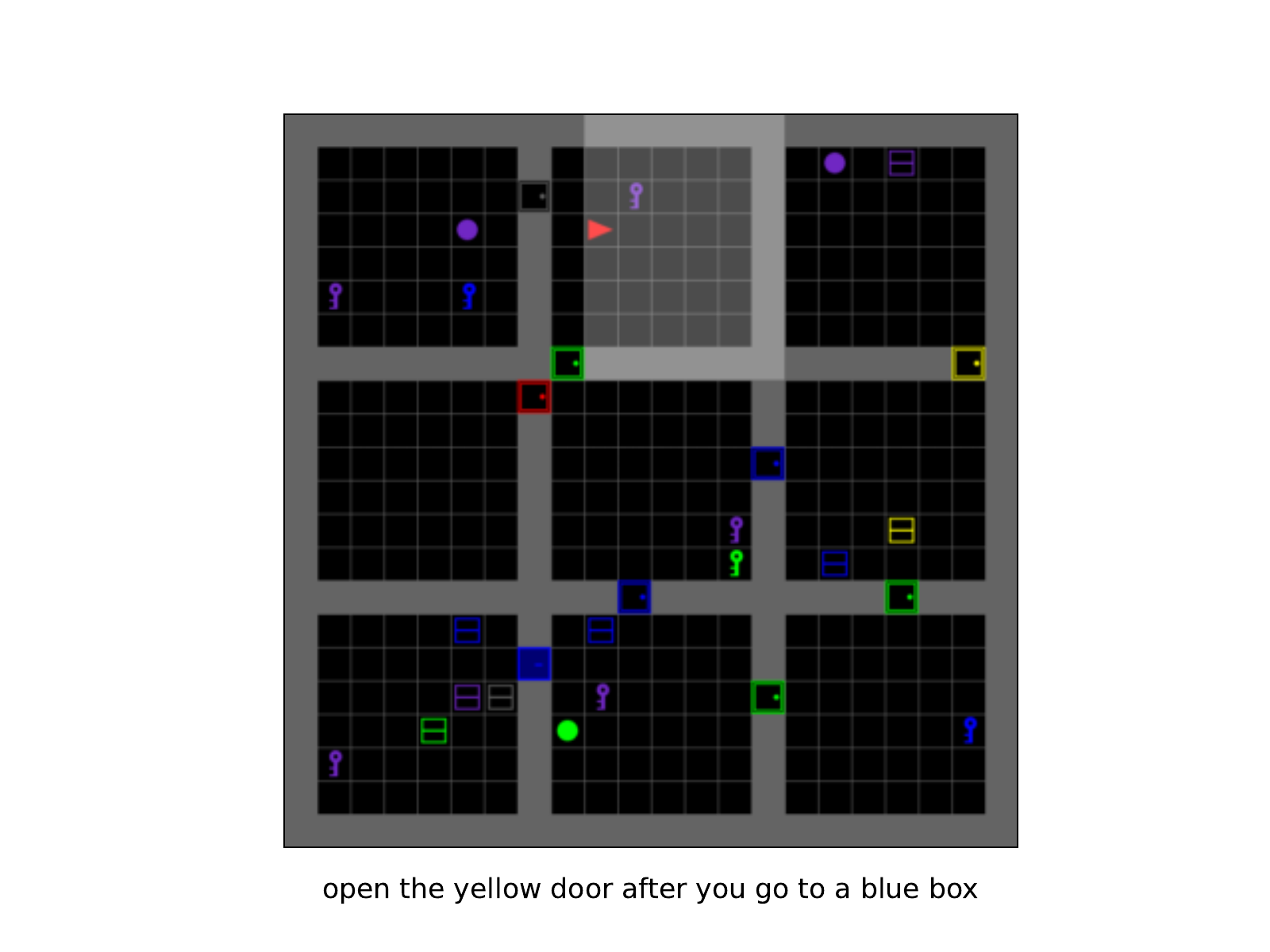}
    \caption{\textbf{BabyAI Demo}. Here we show some examples from the BossLevel environment of BabyAI. The agent is indicated by a red arrow. The bright region in front of the agent shows the partial view of the agent.}
    \label{fig:babyai_demo}
\end{figure}

The BabyAI benchmark \citep{maxime2018babyai} offers a number of gridworld environments in which the agent has to carry out a given instruction. Each environment has 9 rooms arranged in a $3 \times 3$ matrix. Each room has a size of $6 \times 6$. Each environment in BabyAI is partially observable with the agent only having a $7 \times 7$ view of its locality. The total size of the maze is $18 \times 18$. We present a demonstration of some mazes from the BossLevel BabyAI environment in Figure \ref{fig:babyai_demo}. For each BabyAI environment, a new maze is generated for each episode. Each environment in BabyAI tests a different set of competencies of the model. We consider the most difficult environments in the BabyAI benchmark listed below - 
\begin{wrapfigure}{r}{0.45\textwidth}
\vspace{-6mm}
  \begin{center}
    \includegraphics[width=0.43\textwidth]{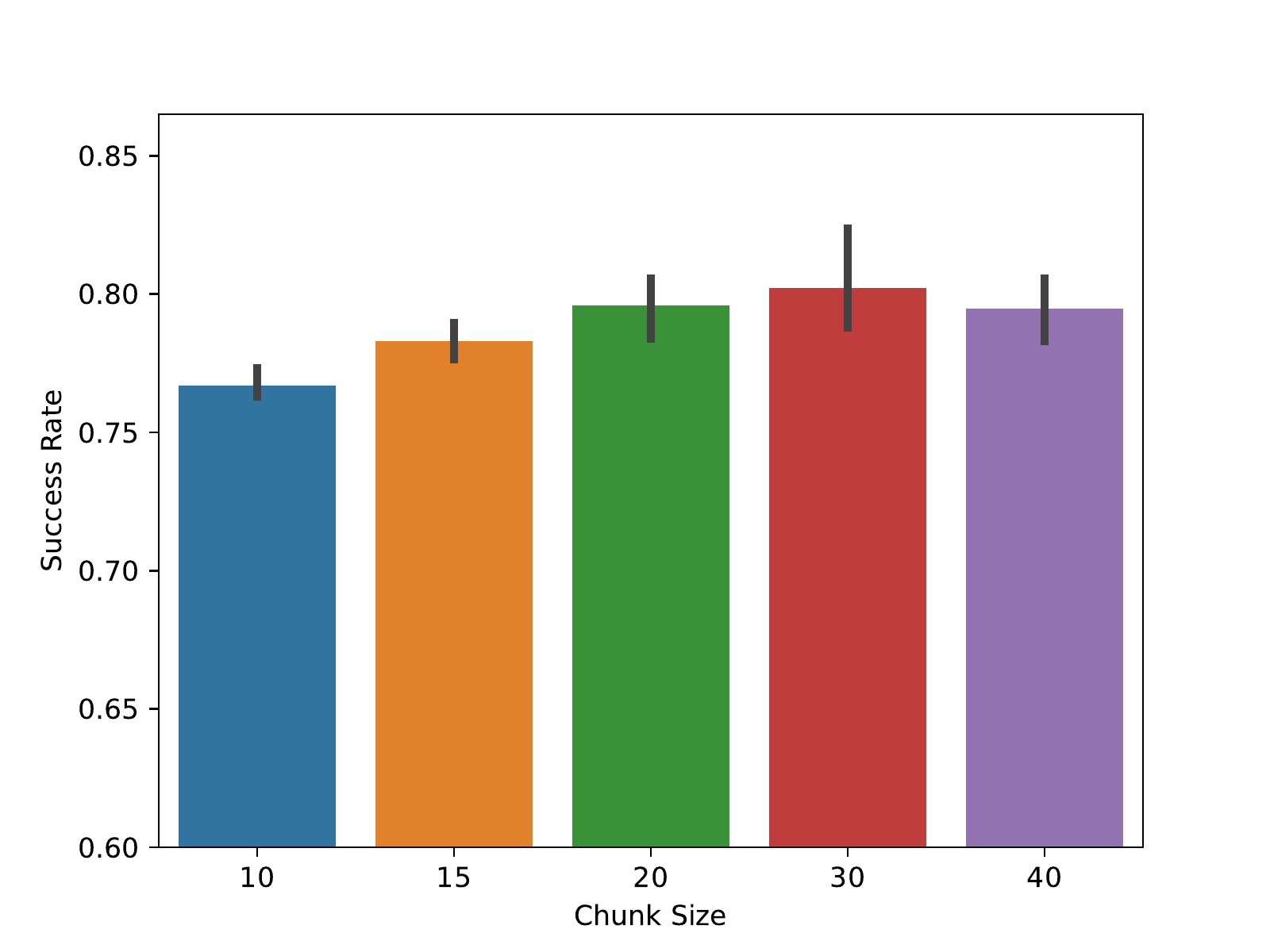}
  \end{center}
  \caption{Here we show the effect of chunk size on the performance of the model for the multi-task BabyAI setting. We can see that similar to Table \ref{tab:listops_chunk_ablation} the model performance hits optimal performance at chunk size 30 above or below which the performance drops.}\label{fig:chunk_size}
  \vspace{-12mm}
\end{wrapfigure}
\begin{itemize}
    \item \textbf{GoToSeq}. A single GoTo instruction tasks the agent to go to a particular location on the grid. GoToSeq consists of a sequence of such GoTo commands.
    \item \textbf{Synth}. This includes a combination of instructions that ask the agent to put one object next to another, go to an object, open a door, or pick up an object.
    \item \textbf{SynthLoc}. Similar to Synth but objects are described using their location rather than appearance.
    \item \textbf{PutNext}. The instructions include tasks to put one object next to another.
    \item \textbf{SynthSeq}. Each instruction is a sequence of commands from the \textit{Synth} environment.
    \item \textbf{BossLevel}. This environment includes instructions that are a combination of all competencies in all other environments of the BabyAI benchmark. Hence, this is the most difficult environment of BabyAI.
\end{itemize}

We train all our models using behavior cloning from an expert policy. We collect 100k expert trajectories from an oracle for each environment. We feed the states for each episode into the model sequentially and task the model to predict the actions at each step. For both Transformer and Transformer + TLB, we use a transformer with 6 layers, embedding dimension set to 512 with 4 heads, FFN dimension set to 1024. For Transformer + TLB, we use 5 temporal latent bottleneck state vectors and chunk size of 30. We perform 1 \textsc{Cross Attention + FFN} per \textsc{Self Attention + FFN}. We train our models for 50000 steps. We evaluate the models by directly deploying them in the environments. We report the success rate across 512 evaluation episodes. An episode is successfully if the agent correctly carries out the given instruction. We train the model using Adam optimizer with a learning rate of 1e-4. Each model is trained on 1 RTX8000 GPU for 24 hours. 

We report the performance of the Transformer baseline and the proposed Transformer + TLB on single tasks in Figure \ref{fig:babyai_single_task}. We can see that in each case, the proposed model converges faster and also outperforms the baseline in some cases.

We also probe the effect of chunk size on the performance of the model. We show the result of this ablation in Figure \ref{fig:chunk_size}. We can see that there is a sweet spot at chunk size 30, above or below which the performance drops. This indicates that a very high chunk size may be too much information for the temporal latent bottleneck state vectors to aggregate while a too low chunk size might lead to large recurrent sequence length which might be difficult to optimize. 

%For the multitask setting reported in Figure \ref{fig:bosslevel}, we consider 50k trajectories per environment. We also report results with 25k trajectories per environment

\iffalse
\begin{figure}
    \centering
    \includegraphics[width = 6cm]{Figures/babyai_multi_task_25000.pdf}
        \includegraphics[width = 6cm]{Figures/babyai_multi_task_50000.pdf}
    %\includegraphics[width = 12cm]{Figures/multiple_task.pdf}
    \caption{\textbf{Multi task Baby AI}. Here we compare the performance of Transformer and Transformer + TB on 8 tasks from the babyai suite of environments for different number of episodes per environment. A single model is trained for all tasks. We can see that Transformer + TB converges and achieves a better performance than Transformer in both the cases.}
    \label{fig:multi_task_babyai}
\end{figure}
\fi
\subsection{Atari} \label{appendix:atari}
For the experiments on Atari, we build on the codebase from \citep{chen2021decision} and extend it by introducing a temporal latent bottleneck. We test it on the same four games (Breakout, Pong, Seaquest, Qbert) as \citep{chen2021decision}. The model is trained on 1\% of the Atari DQN-replay dataset \citep{agarwal2019striving} (500K transitions for each game). 

The models are trained to predict the actions in the offline RL dataset. and are evaluated by directly deploying them in the environments. The results are reported across 10 seeds. The models are trained using a cross-entropy loss for 10 epochs. We use the same hyperparameters as \citep{chen2021decision}. For all the models, we use a transformer with an embedding dimension of 128, 6 layers, 8 heads, and FFN dimension set to 512. The model is trained using AdamW optimizer with a learning rate of 6e-4 with weight decay 0.1.

We train the models on 1 V100 GPU with 32 GB memory for 12 hours.

For the proposed model that incorporates a temporal latent bottleneck the following hyper parameters are used for the temporal latent bottleneck:
\begin{itemize}
    \item \textbf{Pong}: We use chunk size of 18, we set $R = 1$, and use 6 temporal latent bottleneck state vectors.
    \item \textbf{Seaquest}: We use chunk size of 12, we set $R = 2$, and use 6 temporal latent bottleneck state vectors.
    \item \textbf{Qbert}: We use chunk size of 12, we set $R = 2$,  and use 12 temporal latent bottleneck state vectors.
    \item \textbf{Breakout}: We use chunk size of 12, we set $R = 2$,  and use 6 temporal latent bottleneck state vectors.
\end{itemize}
\subsection{Copying Task}\label{sec:copying}
Here we give a detailed description of the copying task and the hyperparameter details of the used models. For a copying task of sequence length 100, the model first receives a sequence of 10 digits between 1 and 8 followed by 100 zeros. The model then receives an indicator input which indicates that the model should start outputting the original sequence it received. The indicator in our case is the digit 9. After receiving the indicator, the model receives 10 more zeros and then it outputs the original sequence again.

For both the Transformer + TLB and the Feedback Transformer model, we use 4 layers and 256 hidden dimension. We use an FFN dimension of 512. For the Transformer + TLB model, we set $R$ to 1. We use a chunk size of 10. We use adam optimizer with learning rate 1e-4. We use a batch size of 100.

%\subsection{Language Modeling} \label{appendix:language_modelling}
%For the language modelling experiments, we use the enwik8 dataset \citep{mahoney2011enwik}. We use a TLB model with 12 layers in the perceptual module with embedding dim set to 1024, FFN dim set to 4096, and 8 heads. We set the chunk size to 512. For the temporal latent bottleneck module, we also use 6 \textsc{Self Attention + FFN} layers per \textsc{Cross Attention + FFN} layer.

\end{document}